\documentclass[runningheads]{llncs}

\usepackage{graphicx}
\usepackage{amsfonts,amssymb}
\usepackage{hyperref}
\usepackage{url}
\usepackage{booktabs}
\usepackage{graphicx}
\usepackage{xspace}
\usepackage{xcolor}
\usepackage{comment}
\usepackage{amsmath}
\usepackage{subcaption}
\usepackage{enumitem}
\usepackage[ruled,vlined,linesnumbered]{algorithm2e}
\usepackage{wrapfig}
\usepackage{cleveref}
\usepackage{ifthen}
\newboolean{shortver}
\setboolean{shortver}{false}

\newcommand{\tool}{\texttt{CoVerD}\xspace}
\newcommand{\cvd}{\texttt{CVD}\xspace}
\newcommand{\cvds}{\texttt{CVDs}\xspace}

\begin{document}
\title{Boosting Few-Pixel Robustness Verification via Covering Verification Designs}
%
%
\author{Yuval Shapira
\and Naor Wiesel
\and Shahar Shabelman
\and Dana Drachsler-Cohen
}
\authorrunning{Y. Shapira et al.}
%
\institute{Technion, Haifa, Israel
\email{\{shapirayuval@campus,wieselnaor@campus,shabelman@campus,ddana@ee\}.technion.ac.il}}

\maketitle              

\renewcommand{\thefootnote}{}
\footnotetext{This preprint has not undergone peer review or any post-submission improvements or corrections. The Version of Record of this contribution is published in Computer Aided Verification: 36th International Conference, CAV 2024, Montreal, QC, Canada, July 24–27, 2024, Proceedings, Part II, and is available online at \url{https://doi.org/10.1007/978-3-031-65630-9_19}.}
\renewcommand{\thefootnote}{\arabic{footnote}}

\begin{abstract}
Proving local robustness is crucial to increase the reliability of neural networks. While many verifiers prove robustness in $L_\infty$ $\epsilon$-balls, very little work deals with robustness verification in $L_0$ $\epsilon$-balls, capturing robustness to few pixel attacks.
This verification introduces a combinatorial challenge, because the space of pixels to perturb is discrete and of exponential size.
A previous work relies on covering designs to identify sets for defining $L_\infty$ neighborhoods, which if proven robust imply that the $L_0$ $\epsilon$-ball is robust. However, the number of neighborhoods to verify remains very high, leading to a high analysis time.
We propose \emph{covering verification designs}, a combinatorial design that tailors effective but analysis-incompatible coverings to $L_0$ robustness verification. 
The challenge is that computing a covering verification design introduces a high time and memory overhead, which is intensified in our setting, where multiple candidate coverings are required to identify how to reduce the overall analysis time.
We introduce \tool, an $L_0$ robustness verifier that selects between different candidate coverings \emph{without constructing them}, but by predicting their block size distribution. 
This prediction relies on a theorem providing closed-form expressions for the mean and variance of this distribution.
\tool constructs the chosen covering verification design \emph{on-the-fly}, while keeping the memory consumption minimal and enabling to parallelize the analysis.
The experimental results show that \tool reduces the verification time on average by up to 5.1x compared to prior work and that  it scales to larger $L_0$ $\epsilon$-balls.
\end{abstract} 
\section{Introduction}\label{sec:intro}
Neural networks are very successful in various applications, most notably in image recognition tasks~\cite{Goodfellow-et-al-2016}. However, neural networks are also vulnerable to adversarial example attacks~\cite{SzegedyZSBEGF13,GrossePM0M16}.
 In an adversarial example attack, an attacker slightly perturbs the input to mislead the network. Many attack models and
 different kinds of perturbations 
 have been considered for neural networks implementing image classifiers~\cite{SzegedyZSBEGF13,GoodfellowSS14,PapernotMJFCS16}. The most commonly studied perturbations are $L_p$ perturbations, where $p$ is $0$~\cite{CroceASF022,YuvilerD23},
 $1$~\cite{Croce021}, $2$~\cite{SzegedyZSBEGF13,Carlini017} or $\infty$~\cite{GoodfellowSS14,Carlini017}. For $L_p$ perturbations, the attacker is given a small budget $\epsilon$ and the goal is to find a perturbed input in the $L_p$ $\epsilon$-ball that causes misclassification. 

\sloppy
In response to adversarial attacks, many verifiers have been proposed to reason about the robustness of neural networks in a given neighborhood of inputs. Most deterministic robustness verifiers analyze robustness in $L_\infty$ $\epsilon$-balls~\cite{Tjeng19,MullerS0PV21,SinghGPV19,GehrMDTCV18,KatzBDJK17,AndersonPDC19}, while 
some deterministic verifiers analyze $L_2$ $\epsilon$-balls~\cite{Leino21,Huang21} or $L_1$ $\epsilon$-balls~\cite{WuZ21,ZhangWCHD18}.
Probabilistic verifiers, often leveraging
randomized smoothing~\cite{CohenRK19}, have been proposed for analyzing $L_{p}$ $\epsilon$-balls for $p\in \{0,1,2,\infty\}$~\cite{LiCWC19,SalmanLRZZBY19,YangDHSR020,DvijothamHBKQGX20}. 
Other verifiers analyze neighborhoods defined by semantic or geometric features (e.g., brightness or rotation)~\cite{KabahaD22,MohapatraWC0D20,BalunovicBSGV19}.
An existing gap is deterministically verifying robustness in $L_0$ $\epsilon$-balls, for a small $\epsilon$, also known as robustness to few pixel attacks. 
In $L_0$ $\epsilon$-balls, $\epsilon$ is the number of pixels that can be perturbed. 
Since $\epsilon$ is an integer (as opposed to 
a real number), we denote it as $t$.
$L_0$ $t$-balls consist of \emph{discrete perturbations}, unlike many other attack models whose perturbations are continuous.
Thus, their analysis is a challenging combinatorial problem. Theoretically, robustness verification of
an $L_0$ $t$-ball can be reduced into a set of robustness verification tasks over $L_\infty$ neighborhoods, each 
allows a specific set of $t$ pixels to be perturbed.
However, this approach is infeasible in practice for $t>2$, since the number of the $L_\infty$ neighborhoods that need to be proven robust is $v\choose t$,
where $v$ is the number of pixels. To illustrate, for MNIST images, where $v=784$, the number of neighborhoods is $1.6\cdot 10^{10}$ for $t=4$, 
$2.4\cdot 10^{12}$ for $t=5$, and 
$3.2\cdot 10^{14}$ for $t=6$. That is, every \emph{minimal} increase of $t$ (by one) increases the neighborhood size by \emph{two orders of magnitude}.

A recent work proposes a deterministic $L_0$ robustness verifier for few pixel attacks, called Calzone~\cite{Shapira23}. Calzone builds on two main observations. 
First, if a network is robust to perturbations of a \emph{specific} set of $k$ pixels, then it is also robust to perturbations of 
any subsumed set of these pixels. 
Second, often $L_\infty$ robustness verifiers can analyze robustness to arbitrary perturbations of $k$ \emph{specific} pixels, 
for values of $k$ that are significantly larger than $t$. 
They thus reduce the problem of verifying robustness in an $L_0$ $t$-ball to 
proving robustness in a set of $L_\infty$ neighborhoods defined by a set of $k$-sized pixel sets, subsuming all possible sets of $t$ pixels. 
To compute the $k$-sized pixel sets, they rely on \emph{covering designs}~\cite{Gordon95newconstructions,todorov1985combinatorial}. 
Given parameters $(v,k,t)$, 
a covering is a set of $k$-sized sets that cover all subsets of size $t$ of a set $[v]=\{1,\ldots,v\}$ (e.g., 
the pixel set).
Covering designs is a field in combinatorics providing construction techniques to compute coverings. The challenge is to compute a covering of minimal size. 
While many covering constructions have been proposed, computing an optimal covering is an open combinatorial problem for most values of $v$, $k$ and $t$. 
Further, most best-known coverings for $t>3$ are far from the best general lower bound, known as the Sch\"onheim bound~\cite{schonheim}.
This severely impacts the analysis time of Calzone.
In practice, Calzone often does not complete within the five hour timeout when analyzing $L_0$ $5$-balls.
To scale, it is crucial to lower the number of analyzed sets.
While there are effective covering constructions renowned for the small coverings they compute, they are limited to specific
values of $v$ and $k$, which are incompatible for the analysis of $L_0$ robustness. 
Since Calzone treats covering constructions as black-box, it is limited to rely on analysis-compatible 
coverings and cannot benefit from these effective constructions.

To boost the robustness verification of few pixel attacks, we propose 
a new covering type, called a \emph{covering verification design} (\cvd), tailoring covering designs for $L_0$ robustness verification.
\cvd relies on a highly effective construction to obtain an analysis-incompatible covering and \emph{partially induces} it 
to an analysis-compatible covering, where sets can have different sizes.
   Although the exact sets and their sizes depend on a random choice, 
we prove that the mean and variance of the set sizes are independent of this choice and have closed-form expressions.
Partially inducing this effective construction has been proposed before~\cite{Ray-Chaudhuri1968}, 
however it has been proposed for another combinatorial design, requiring a bound on the maximal set size in the covering, unlike \cvd. 
We demonstrate that the sizes of \cvds are \emph{lower} by 8\%
for $t=4$ and by 15\% for $t=5$ than the respective Sch\"onheim 
lower bound. This improvement, enabled by considering a new type of coverings, is remarkable for scaling $L_0$ robustness analysis. To date, 
 for analysis-compatible values of $v$ and $k$ and for $t \geq3$, it is impossible to obtain an optimal covering design, and even if we obtained it,
 its size is \emph{at least} the  
 Sch\"onheim bound. In particular, Calzone's considered coverings are larger by 4x than the Sch\"onheim 
lower bound for $t=4$ and by 8.4x for $t=5$.
 While promising, \cvds raise a practical challenge: their construction as well as their final size introduce 
 a high memory overhead.
Further, to minimize the analysis time, the verifier chooses between \emph{multiple} coverings. 
However, the total memory overhead makes it infeasible
 to store these coverings in a covering database without limiting their size  
 (like Calzone does). 

We introduce \tool, an $L_0$ robustness verifier, boosting Calzone's performance by leveraging \cvds. 
\tool has two main components, \emph{planning} and \emph{analysis}.
The planning component predicts the \cvd that will allow it to minimize the overall analysis time. 
To reduce the memory overhead, it predicts the best \cvd out of many candidates, 
\emph{without constructing the candidates}. This prediction relies on estimating the set size distribution of a candidate covering, using our expressions for the mean and variance.
The analysis component constructs the chosen \cvd.
The challenge is that the original covering that is being induced may be too large to fit the memory.
To cope, \tool induces the covering while constructing the original covering. Further, it constructs 
\emph{on-the-fly} a partitioning of the \cvd so that the analysis can be parallelized over multiple GPUs. 
Another advantage of the on-the-fly construction is that \tool does not need to prepare coverings for every image dimension in advance.
This both saves memory consumption and makes \tool suitable for \emph{any} image classifier, without requiring to precompute
coverings for new image dimensions, as Calzone requires.

We evaluate \tool on convolutional and fully-connected networks, trained for MNIST, Fashion-MNIST, and CIFAR-10.
\tool is faster than Calzone in verifying robust $t$-balls on average by 2.8x for $t=4$ and by 5.1x for $t=5$.
Further, \tool scales to more challenging
$t$-balls than Calzone. In particular, it verifies some $6$-balls, which Calzone does not consider at all, within $42$ minutes. 

\section{Background}\label{sec:background}
In this section, we define the problem of verifying robustness of an image classifier in an $L_0$ $t$-ball and provide background on Calzone~\cite{Shapira23}. 

 \paragraph{$L_0$ robustness verification} 
 We address the problem of determining the local robustness of an image classifier in an $L_0$ $t$-ball of an image $x$.
 An image classifier $N$ takes as input an image $x$ consisting of $v$ pixels, each ranges over $[0,1]$ (all definitions extend to colored images, but omitted for simplicity's sake). It returns a vector consisting of a score for every possible class. The classification the classifier $N$ assigns to an input image $x$ is the class with the maximal score: $c_x=\texttt{argmax}(N(x))$. We focus on classifiers implemented by neural networks. Specifically, our focus is on fully-connected and convolutional networks, since many $L_\infty$ robustness verifiers can analyze them~\cite{SinghGPV19,GehrMDTCV18,KatzBDJK17,AndersonPDC19,Tjeng19,MullerS0PV21}. However, like Calzone, \tool is not coupled to the underlying implementation of the classifier and can reason about any classifier for which there are $L_\infty$ robustness verifiers that it can rely on. The problem we study is determining whether a classifier $N$ is locally robust in the $L_0$ $t$-ball of an input $x$, for $t\geq 2$. That is, whether every input whose $L_0$ distance from $x$ is at most $t$ is classified by $N$ as $x$ is classified. Formally, the $t$-ball of $x$ is $B_t(x)=\{x' \mid ||x'-x||_0 \leq t\}$ and $N$ is locally robust in $B_t(x)$ if $\forall x'\in B_t(x).\ \texttt{argmax}(N(x'))=\texttt{argmax}(N(x))$. We note that the $L_0$ distance of two images is the number of pixels that the images differ, that is $||x'-x||_0 = |\{i \in [v]\mid x_i \neq x_i'\}|$ (where $[v]=\{1,\ldots,v\}$). In other words, an $L_0$ perturbation to an image $x$ can arbitrarily perturb up to $t$ pixels in $x$.

\begin{figure*}[t]
    \centering
  \includegraphics[width=1\linewidth, trim=0 90 0 0, clip, page=2]{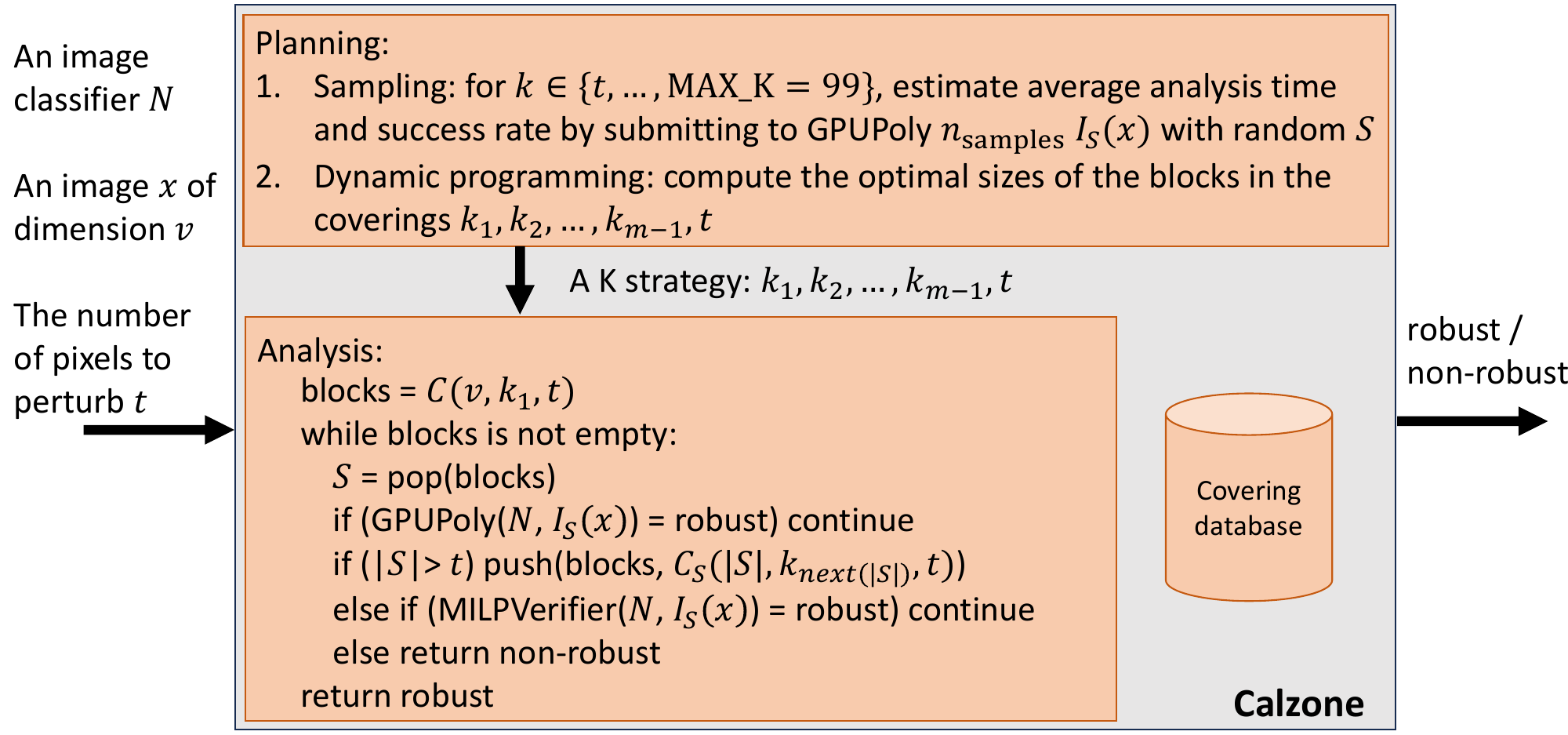}
    \caption{The Calzone $L_0$ robustness verifier.}
    \label{fig:background}
\end{figure*}

 \paragraph{Calzone} Calzone, depicted in~\Cref{fig:background}, is an $L_0$ robustness verifier. It verifies by determining the robustness of a classifier $N$ in all neighborhoods in which a specific set of pixels $S$ is arbitrarily perturbed, for every $S \subseteq [v]$ of size $t$. Namely, to prove robustness, it
 has to determine for every such $S$ whether $N$ classifies the same all inputs in the neighborhood consisting of all images that are identical to $x$ in all pixels, but the pixels in $S$. We denote this neighborhood by $I_S(x)=\{x' \in [0,1]^v \mid \forall i\notin S.\ x'_i=x_i\}$. Such neighborhoods can be specified as a sequence of intervals, one for every pixel, where the $i^\text{th}$ interval is $[0,1]$ if $i\in S$ (i.e., it can be perturbed) or $[x_i,x_i]$ if $i\notin S$ (i.e., it cannot be perturbed). Most existing $L_\infty$ robustness verifiers can determine the robustness of such interval neighborhoods. However, verifying $v\choose t$ interval neighborhoods, one for every selection of $t$ pixels to perturb, is practically infeasible for $t>2$. Instead, Calzone builds on the following observation: if $N$ is locally robust in a neighborhood $I_{S'}(x)$ for $S'\subseteq [v]$ of size $k>t$, then $N$ is also robust in every $I_{S}(x)$, for $S\subseteq S'$ of size $t$.
    This observation allows Calzone to leverage \emph{covering designs} to reduce the number of neighborhoods analyzed by an $L_\infty$ verifier. 
    Given three numbers $(v,k,t)$, for $t\leq k\leq v$, a covering $C(v,k,t)$ is a set of \emph{blocks}, where (1)~each block is subset of size $k$ of $[v]$ and (2)~the blocks cover all subsets of $[v]$ of size $t$: for every $S\subseteq [v]$ of size $t$, there is a block $B\in C(v,k,t)$ such that $S\subseteq B$. Coverings are evaluated by their size, $|C(v,k,t)|$, where the smaller the better.
    We next describe the components of Calzone: analysis, planning and covering database.

     \paragraph{Calzone's analysis} 
     Calzone begins the analysis by obtaining a covering $C(v,k_1,t)$ from its covering database, where $k_1$ is determined by the planning component (described shortly). It pushes all blocks in the covering into a stack. It then iteratively pops a block $S$ from the stack and verifies the robustness of $N$ in $I_S(x)$ by running GPUPoly~\cite{MullerS0PV21}. 
  GPUPoly is a sound $L_\infty$ robustness verifier which is highly scalable because it performs the analysis on a GPU. However, it relies on a linear relaxation and thus may fail proving robustness due to overapproximation errors.
  If it determines that $I_S(x)$ is robust, Calzone continues to the next block. Otherwise, Calzone performs an exact analysis or refines the block. If $|S|= t$, Calzone invokes a sound and complete mixed-integer linear programming (MILP) verifier~\cite{Tjeng19}. If it determines that $I_S(x)$ is not robust, Calzone returns \emph{non-robust}, otherwise Calzone continues to the next block. If $|S|$ is greater than $t$, Calzone refines $S$ by pushing to the stack all blocks in 
  a covering for $S$ and $t$. The blocks' size is $k_{i+1}$, which is the block size following the current block size $k_i=|S|$, as determined by the planning component. The covering is obtained by retrieving from the covering database the covering $C(|S|,k_{i+1},t)$ and renaming the numbers in the blocks 
  to range over the numbers in $S$ (instead of $[|S|]$), denoted as $C_S(|S|,k_{i+1},t)$.
  If Calzone observes an empty stack, it returns \emph{robust}. This analysis is proven sound and complete. To scale, Calzone parallelizes the analysis over multiple GPUs (for GPUPoly) and CPUs (for the MILP verifier). Technically, the first covering is split between the GPUs, each independently analyzes its assigned blocks and refines if needed. 
  
  \paragraph{Calzone's planning} The planning determines the block sizes of the first covering and of the refinements' coverings. These are given as a \emph{K strategy}, a decreasing series $k_1>\ldots > k_m$, where $k_1\leq \text{MAX\_K}=99$ and $k_m=t$. Calzone predicts the K strategy that minimizes the overall analysis time using dynamic programming, defined over the analysis time of the first covering, the average fraction of blocks that will be refined, and the analysis time of the refined blocks.  
  This computation requires GPUPoly's success rate and average analysis time for neighborhoods $I_S(x)$, for all $|S|\leq \text{MAX\_K}$. These are estimated by sampling $n_\text{samples}=400$ sets $S$ for every $k\leq \text{MAX\_K}$ and submitting their neighborhood $I_S(x)$ to GPUPoly.   
 
 \paragraph{Calzone's covering database} As described, the analysis obtains coverings from a database. This database has been populated  
 by obtaining well-optimized coverings from online resources and extending them for large values of $v$ and $k$ using general covering constructions.  
 Because of these general constructions, the database's coverings tend to be far from the Sch\"onheim bound~\cite{schonheim}, the best-known general lower bound, especially for large values of $v$ (the image dimension). This inefficiency results in longer analysis, since more blocks are analyzed.

\section{Our Approach: Covering Verification Designs}\label{sec:approach}

To scale Calzone's analysis, it is crucial to reduce the number of blocks that are analyzed by GPUPoly or the MILP verifier.
A dominant contributor to this number is 
the size of the first covering, for two reasons. First, the first covering is over a large $v$ (the image dimension), thus its size is significantly larger than the sizes of coverings added upon refinement, which are over significantly smaller $v$ 
(typically $v\leq 80$ and at most $v\leq \text{MAX\_K}$). Second, 
the first covering has an accumulative effect on the number of refinements, and consequently it dominates the analysis time.
Reducing this size is theoretically possible by relying on \emph{finite geometry covering constructions}~\cite{Ray-Chaudhuri1968,Abraham1968,Gordon95newconstructions}, which are renowned for computing very small coverings. However, finite geometry coverings are limited to $(v,k,t)$ triples in which $v$ and $k$ are defined by related mathematical expressions over~$t$.
In Calzone's analysis, the first covering has to be defined over a given $v$ (the image dimension) and $t$ (the number of perturbed pixels). Thus, for some values of $v$ and $t$, there is no finite geometry covering. For the other values, there are very few values for $k$, leading to long analysis either because they are large and have a low success rate, triggering many refinements, or small and have very large coverings.  
We propose to tailor \emph{induced coverings} for $L_0$ robustness analysis in order to leverage finite geometry coverings. To this end, we introduce a new type of a covering design, called a \emph{covering verification design} (\cvd). We next provide background on finite geometry coverings and induced coverings. We then define \emph{partially-induced coverings} and our new covering type.
We discuss its properties, its effectiveness, and the practical challenges in integrating it to $L_0$ verification.

\begin{wrapfigure}[8]{r}{0.35\linewidth}
\vspace{-0.06cm}
\hspace{0.5cm}
  \includegraphics[trim=0 125 0 0, clip, width=0.7\linewidth]{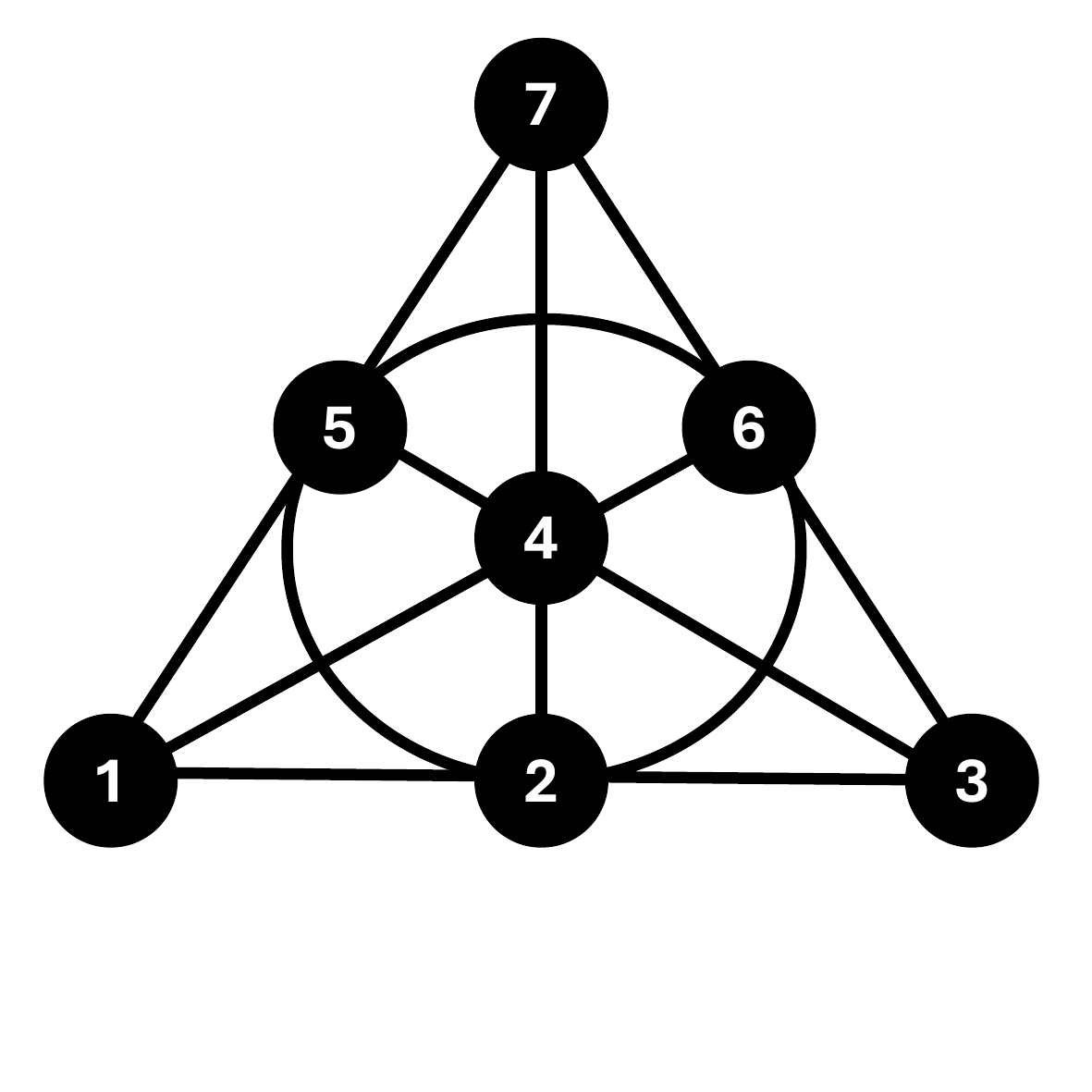}
  \caption{The Fano Plane.}
  \label{fig:fano}
  \end{wrapfigure}
\paragraph{Finite geometry coverings}
Finite geometry covering constructions are widely known for obtaining small (sometimes optimal) coverings~\cite{Ray-Chaudhuri1968,Abraham1968,Gordon95newconstructions}.
Popular finite geometry constructions rely on projective geometry (PG) or affine geometry (AG).
We focus on PG, but our approach extends to AG.
A PG construction views the problem of constructing a covering for a given ($v$, $k$, $t$) from a finite geometry point of view,  where $v$ is the number of points in the geometry.
 It constructs coverings by computing flats (linear subspaces) of dimension $t-1$, each containing $k$ points. 
Since every $t$ points from $[v]$ are contained in at least one flat~\cite{Ray-Chaudhuri1968}, the flats provide a covering.
\Cref{fig:fano} shows \emph{the Fano plane}, a well-known example. 
In this example, there are $v=7$ points, 
the flats are of dimension $t-1=1$ (the lines and the circle), each containing $k=3$ points.  
The set of flats forms a covering, where each flat is a block: 
$C(7,3,2)=\{\{1,2,3\},\{1,4,6\},\{1,5,7\},\{2,4,7\},\{2,5,6\},\{3,4,5\},\{3,6,7\} \}$. 
PG coverings exist for triples where $v=\frac{q^{m+1}-1}{q-1}$ and $k=\frac{q^t-1}{q-1}$, for a prime power $q$ and $m\geq t\geq 2$ (it also exists for $m=t-1$, but then $v=k$, which is unhelpful to our analysis).  
Because PG is restricted to such triples, Calzone cannot effectively leverage it for the first covering, whose $v$ and $t$ are given.  
This is because for common image dimensions (e.g., $v=784$ for MNIST and $v=1024$ for CIFAR-10), there are no suitable $q$ and $m$.
Even if there are suitable $q$ and $m$, there are very few possible $k$ values, which are unlikely to include or be close to an optimal value of $k$. 
Thus, either they are smaller than an optimal $k$, leading to larger coverings and a longer analysis time, or that they are larger than an optimal $k$ and have a lower success rate, leading to many refinements, resulting, again, in a longer analysis time. 
For example, for $v=364$ and $t=5$, the only suitable values are $q=3$ and $m=5$ (i.e., $364=({3^{5+1}-1})/({3-1})$), namely there is only one triple for these values of $v$ and $t$. In this triple, 
$k=({3^5-1})/({3-1})=121$. Since $k\approx\frac{v}{3}$, neighborhoods $I_S(x)$ for which $|S|=121$ are not likely to be robust, thus such $k$ is likely to have a low success rate. 
\emph{Induced coverings}~\cite{Gordon95newconstructions} enable to leverage finite geometry coverings for other $(v,k,t)$ triples, as next explained.

\paragraph{Induced coverings} 
Given $v\leq v'$ and $k\leq k'$,
a covering $C(v',k',t)$ can be \emph{induced} to form a covering $C(v,k,t)$~\cite{Gordon95newconstructions}.
The induced covering is obtained in three steps. First, we select a subset of numbers of size $v$, denoted $L\subseteq [v']$, and remove every $l\in [v']\setminus L$ from every block in $C(v',k',t')$. This results in a set of blocks of different sizes that covers all subsets of $L$ of size $t$~\cite[Lemma 1]{Ray-Chaudhuri1968}. This follows since every subset $S\subseteq L$ of size $t$ is a subset of $[v']$ and thus there is $B\in C(v',k',t)$ such that $S\subseteq B$. The first step removes from $B$ only numbers from $[v']\setminus L$ and thus $S$ is contained in the respective block to $B$ after this step. 
The next two steps fix blocks whose size is not $k$.
The second step extends every block whose size is smaller than $k$ with numbers from $L$. The third step refines every block whose size is larger than $k$ to multiple blocks of size $k$ that cover all of its subsets of size $t$. This step significantly increases the number of blocks, unless the number of blocks larger than $k$ is negligible.
We note that these steps provide a covering over the numbers in $L$ (i.e., $C_{L}(|L|,k,t)$). A covering for $(|L|,k,t)$ can be obtained by renaming the numbers to range over $[|L|]$.

\paragraph{Partially-induced covering} 
Our new covering design is an instance of a \emph{partially-induced covering}. A partially-induced covering is the set of blocks obtained by the first step, where the blocks 
cover all subsets of $L$ of size $t$ and 
are of different sizes. 
For example, for the Fano plane  
and $L_1=\{4,5,6,7\}$, the partially-induced covering is: 
$C_1 = \{\{\}, \{4,6\}, \{5,7\}, \{4,7\}, \{5,6\}, \{4,5\}, \{6,7\}\}$, while for $L_2=\{1,2,3,4\}$, it is: 
$C_2 = \{\{1,2,3\}, \{1,4\}, \{1\}, \{2,4\}, \{2\}, \{3,4\}, \{3\}\} $. 
Partially-induced coverings have two benefits in our setting: (1)~by not extending blocks whose size is smaller than $k$, we increase the likelihood
that GPUPoly will prove their robustness, and (2)~by not refining blocks whose size is larger than $k$, we 
(a)~preserve the number of blocks as in the original covering, (b)~provide GPUPoly an opportunity to verify these blocks, and (c)~rely on the optimal refinement sizes (computed by the dynamic programming) for blocks that GPUPoly fails proving robustness.
Our covering design 
partially induces PG coverings, to obtain additional benefits for $L_0$ robustness verification.  

\paragraph{Covering verification designs}
Given the number of pixels $v$ and the number of pixels to perturb $t$, a covering verification design (\cvd) is the set of blocks obtained by partially inducing a PG covering $C(v',k',t)$, where $v\leq v'$, using a random set of numbers $L\subseteq[v']$ of size $v$. The numbers in the blocks can later be renamed to range over $[v]$. 
For example, since the Fano plane is a PG covering, the partially-induced coverings $C_1$ and $C_2$ are \cvds.
A \cvd has two important properties. First, it is a partially-induced covering and thus 
has all the aforementioned advantages in our setting. In particular, 
its size is equal to the size of the original covering, which is highly beneficial since \cvd induces from PG coverings, known for their small size. 
Second, although different sets $L$ lead to different block size distributions, we prove that the
mean block size and its variance are \emph{the same} regardless of the choice of $L$. Further, we identify closed-form expressions for them and 
show that the variance is bounded by the mean. 
For example, although the block size distributions of $C_1$ and $C_2$ are different, they have the same average block size ($\frac{12}{7}$) and the same variance ($\frac{24}{49}$). 
This property has practical advantages: (1)~it allows us to estimate the block size distribution (\Cref{sec:CoveringDistributionEstimator}), and (2)~since the variance is bounded by the mean, the smaller the mean block size, the less likely that there are very large blocks, 
which are less likely to be proven robust by GPUPoly.
To prove this property, we rely on the fact that PG coverings (and AG coverings) are also a combinatorial design called a \emph{balanced incomplete block design} (BIBD)~\cite[Part VII, Proposition 2.36]{alma990022986490203971}. We next describe BIBD and then state our theorem on its mean and variance.

\paragraph{BIBD}
Given positive integers $(v,b,r,k,\lambda)$, a BIBD is a set of $b$ blocks, each is a subset of $[v]$ of size $k$, such that every $i \in [v]$ appears in $r$ blocks and every $i\neq j \in [v]$ appear together in $\lambda$ blocks. 
For example, the Fano plane is a BIBD with $v=7,b=7,r=3,k=3,\lambda=1$. This is because it has $b=7$ blocks, each block is a subset of $[v]=\{1,\dots,7\}$ of size $k=3$, every number in $\{1,\dots,7\}$ appears in $r=3$ blocks and every two different numbers appear together in $\lambda =1$ block. 
Given a BIBD with parameters $(v',b,r,k',\lambda)$, we define a partially-induced BIBD for $v\leq v'$ by 
selecting a subset of numbers $L\subseteq [v']$ of size $v$ and removing every $l\in [v']\setminus L$ from every block in the BIBD (empty blocks or repetitive blocks are kept).
While the distribution of the induced blocks' sizes depends on $L$, the mean block size and its variance depend only on $v,v',k'$. 

\begin{theorem}\label{mean_variance_thm}
Given a $(v',b,r,k',\lambda)$-BIBD, for $v'>1$, and $1\leq v\leq v'$, for every $L\subseteq [v']$ of size $v$, 
the mean $\mu_{v',k',v}$ and variance $\sigma_{v',k',v}^2$ of the block sizes in the partially-induced BIBD satisfy:
\begin{enumerate}[nosep,nolistsep]
    \item 
    $\mu_{v',k',v} = \frac{vk'}{v'}$
    \item 
    $\sigma_{v',k',v}^2 = \mu_{v',k',v} \left (1+\frac{(v-1)(k'-1)}{v'-1}-\mu_{v',k',v}\right )=
    \frac{vk'}{v'} \left (1+\frac{(v-1)(k'-1)}{v'-1}-\frac{vk'}{v'}\right )$
    \item $\sigma_{v',k',v}^2\leq \mu_{v',k',v}$
\end{enumerate}   
\end{theorem}
\newcounter{rtaskno}
\newcommand{\rtask}[1]{\refstepcounter{rtaskno}\label{#1}}

\begin{proof}
    1. We prove $\mu_{v',k',v} = \frac{vk'}{v'}$.
        Since $|L|=v$ and $r$ is the number of occurrences of every number in all blocks, the sum of the sizes of the induced blocks is $vr$. 
        By counting arguments, for a BIBD it holds that $rv' = bk'$~\cite[Part II, Proposition 1.2]{alma990022986490203971}, and so $r=\frac{bk'}{v'}$. 
        That is, the sum of the induced blocks' sizes is $\frac{vbk'}{v'}$. The mean is obtained by dividing by the number of blocks $b$: $\mu_{v',k',v}=\frac{vk'}{v'}$.
            
2. 
We prove $\sigma_{v',k',v}^2 = \mu_{v',k',v} \left (1+\frac{(v-1)(k'-1)}{v'-1}-\mu_{v',k',v}\right )$.\\
Let $Z\in {\mathbb{N}_0}^b$ be a vector such that, for every $n\in [b]$, $Z_n$ is the size of block $n$ in the partially-induced BIBD.
It can be written as $Z = A^Tx_L$, where $A$ represents the BIBD and $x_L$ the set $L$, used for partially inducing the BIBD. 
The matrix $A$ is a $v'\times b$ incidence matrix, where $A[m,n] = 1$ if $m$ is in block $n$
  and $A[m,n]=0$ otherwise. 
  The vector $x_L$ is a $v'$-dimensional vector, where $x_L[m]=1$ if $m\in L$ and $x_L[m]=0$ otherwise.
Thus, the average of the squares of the block sizes, denoted $\mathbb{E}[Z^2]$, is $\mathbb{E}[Z^2]= \frac{1}{b}\left(\sum_{n=1}^{b}(A^Tx_L)_n^2\right) = 
\rtask{myeq}\frac{1}{b}\Vert A^Tx_L\Vert_2^2$ (\ref{myeq}).\\
By the variance definition, $\sigma_{v',k',v}^2=\mathbb{E}[Z^2] - \mu_{v',k',v}^2$. 
Thus, we need to show: $\mathbb{E}[Z^2]=\mu_{v',k',v} (1+\frac{(v-1)(k'-1)}{v'-1})=
\frac{vk'}{v'}(1+\frac{(v-1)(k'-1)}{v'-1})=\frac{k'}{v'}v+\frac{k'(k'-1)}{v'(v'-1)}v(v-1)
$.
By counting arguments~\cite{alma990022986490203971}, we have $\frac{k'}{v'}=\frac{r}{b}$ and 
$\frac{k'(k'-1)}{v'(v'-1)}=\frac{\lambda}{b}$.
Namely, it suffices to show: 
$\mathbb{E}[Z^2]=\frac{1}{b}\left (rv + \lambda v(v-1)\right )$. 
By~(\ref{myeq}), we can show $\Vert A^Tx_L\Vert_2^2 = r v + \lambda v(v-1)$.
We prove by induction on $v=|L|$ that $\Vert A^Tx_L\Vert_2^2 = r v + \lambda v(v-1)$:\\
{\bf Base} For $v=1$, we show $\Vert A^Tx_L\Vert_2^2 = r \cdot 1+ \lambda \cdot 1\cdot 0$: Since $v=|L|=1$, 
by definition of a BIBD, the vector of the induced blocks' sizes $Z$ has $r$ ones and the rest are zeros. 
Thus,  $\Vert Z\Vert_2^2=r$. Since $Z=A^Tx_L$, the claim follows.\\
{\bf Induction hypothesis} Assume that the claim holds for every $1,\ldots,v$. \\
{\bf Step} Let $L\subseteq [v']$ such that $|L|=v+1$. Pick some $i\in L$ and define $L'=L\setminus \{i\}$ of size $v$.
We get $x_{L}=x_{L'}+e_i$, where $e_i$ is the $i^\text{th}$ standard unit vector. Thus:
$$\Vert A^Tx_{L}\Vert_2^2 = \Vert A^T(x_{L'}+e_i)\Vert_2^2 = \Vert A^Tx_{L'}\Vert_2^2 + \Vert A^Te_i\Vert_2^2 + 2\left\langle A^Tx_{L'},A^Te_i\right\rangle$$
\begin{itemize}[nosep,nolistsep]
  \item By the induction hypothesis, $\Vert A^Tx_{L'}\Vert_2^2= rv + \lambda v(v-1)$.
  \item Since $e_i$ can be viewed as $x_{L''}$ for some $L''$ of size 1, we get
  $\Vert A^Te_{i}\Vert_2^2= r$.
  \item We show $\left\langle A^Tx_{L'},A^Te_i\right \rangle = x_{L'}^T\left (AA^T\right) e_i=\lambda v$:
Since $A$ is an incidence matrix of a BIBD, 
$AA^T$ is the matrix with $r$ on the diagonal and $\lambda$ elsewhere~\cite[Part II, Theorem 1.8]{alma990022986490203971}. 
Therefore, $\left (AA^T\right) e_i$ is a vector whose entries are $\lambda$ except for the $i^\text{th}$ entry which is $r$. 
The vector $x_{L'}$ has $v$ ones and $0$ on the $i^\text{th}$ entry (since $i\notin L'$).
Thus, their dot product is $x_{L'}^T\left (AA^T\right) e_i = \lambda v$.
\end{itemize}
   Putting it all together:
$\Vert A^Tx_{L}\Vert_2^2 = rv + \lambda v(v-1) + r +2\lambda v = r(v+1) + \lambda (v+1)v$.

3. We show $\sigma_{v',k',v}^2 \leq \mu_{v',k',v}$ by showing that $1+\frac{(v-1)(k'-1)}{v'-1}-\mu_{v',k',v}\leq 1$.
Since $\mu_{v',k',v}=\frac{vk'}{v'}$, we show $\frac{(v-1)(k'-1)}{v'-1}\leq\frac{vk'}{v'}$.
We have $1 \leq v \leq v'$ and $1<v'$, thus we get $\frac{v-1}{v'-1}\leq\frac{v}{v'}$.
Since $k'-1\geq 0$, we get $\frac{(v-1)(k'-1)}{v'-1}\leq \frac{v(k'-1)}{v'} \leq \frac{vk'}{v'}$. 
\qed        
\end{proof}

 \begin{figure*}[t]
 \centering
\begin{subfigure}{.5\textwidth}
  \centering
  \includegraphics[height=4cm,width=1\linewidth]{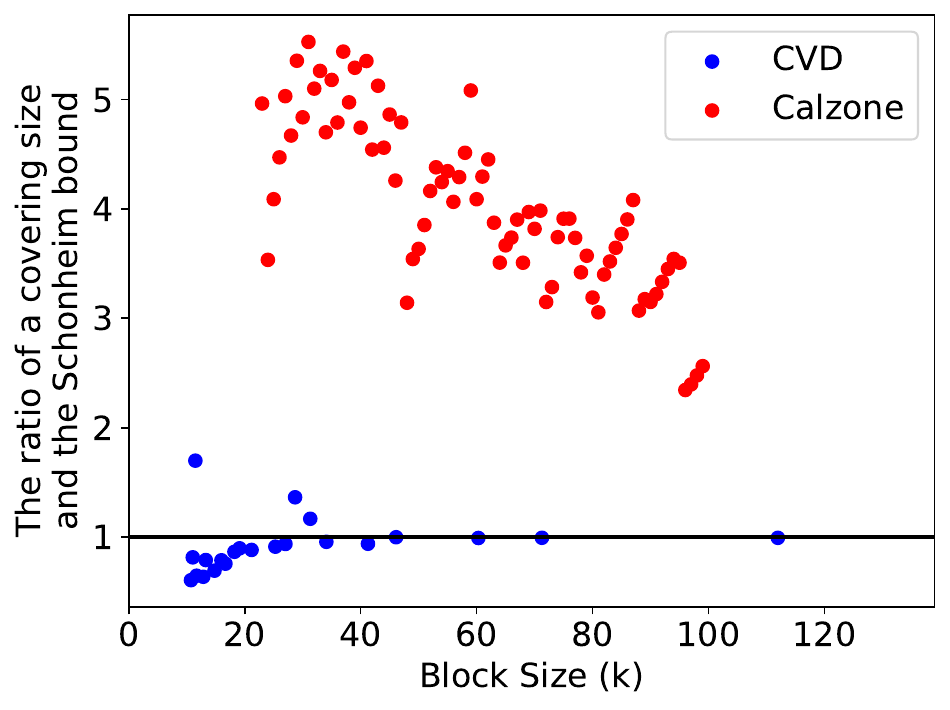}
  \caption{The plot for $t=4$.}
  \label{fig:sub1}
\end{subfigure}%
\begin{subfigure}{.5\textwidth}
  \centering
  \includegraphics[height=4cm,width=1\linewidth]{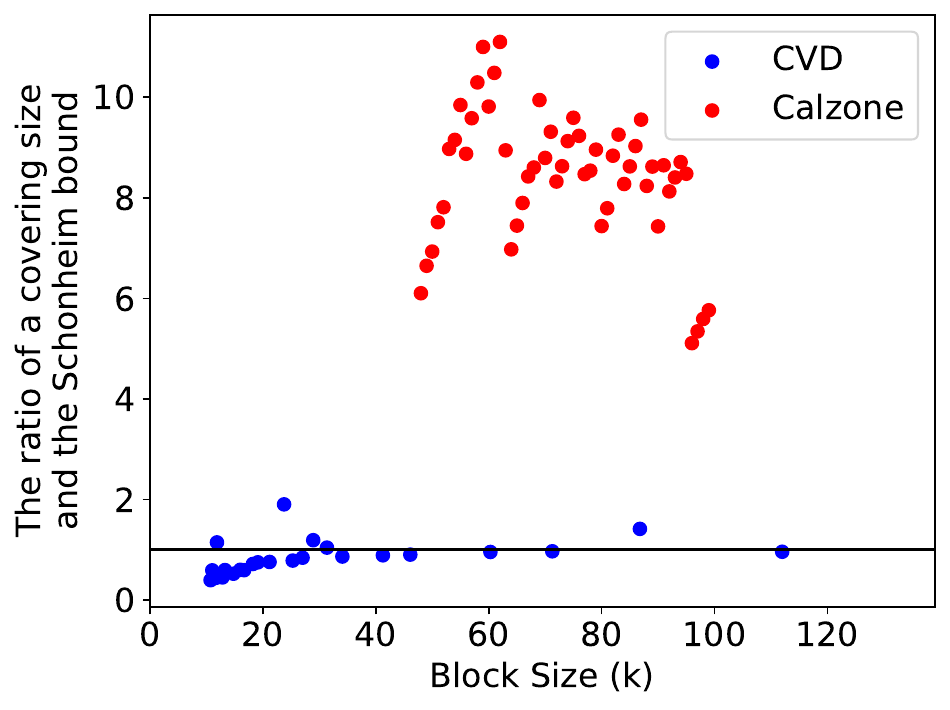}
  \caption{The plot for $t=5$.}
    \label{fig:sub2}
  \end{subfigure}
\caption{The ratio of \cvd sizes and their respective Sch\"onheim bound vs. the ratio of Calzone's covering sizes and their Sch\"onheim bound. The black line is ratio $1$, i.e., coverings whose sizes are equal to the respective Sch\"onheim bound.}
\label{fig:histcovs}
\end{figure*}

\paragraph{Size of covering verification designs} \cvds enable us to obtain coverings whose sizes are small, often close or better than their respective Sch\"onheim bound. 
Given a \cvd whose mean block size is a real number $k$, we define its 
respective Sch\"onheim bound as the bound for the covering design of $(v,\lceil{k} \rceil,t)$. 
Note that this bound is not a lower bound on the size of the \cvd, since the \cvd can have blocks larger than 
$\lceil{k} \rceil$ and thereby be smaller than 
covering designs for $(v,\lceil{k} \rceil,t)$.
Still, comparing to this bound enables understanding how much smaller our coverings are compared to the coverings considered by Calzone, whose sizes are lower bounded by the Sch\"onheim bound.
\Cref{fig:histcovs} shows the ratio of the sizes of our \cvds and their respective Sch\"onheim bound and the ratio of the sizes of Calzone's covering designs and their Sch\"onheim bound.
The comparison is for $v=784$ and $t=4$ (\Cref{fig:sub1}) and $t=5$ (\Cref{fig:sub2}). We compute \cvds from different PG coverings and the figure shows \cvds whose mean block size $k$ is at least $10$. For Calzone, we show all coverings in its database.
The plots demonstrate that typically the size of a \cvd is smaller or equal to the Sch\"onheim bound, and on average, the ratio is $0.92$ for $t=4$ and $0.85$ for $t=5$. In contrast, Calzone's coverings are significantly larger than the Sch\"onheim bound, on average the ratio is $4.04$ for $t=4$ and $8.44$ for $t=5$. The plots also show that Calzone has many more coverings than the number of \cvds. This is because Calzone relies on general techniques to compute coverings and thus it can generate a covering for every $k\leq \text{MAX\_K}=99$
(except that it is limited to coverings with at most $10^7$ blocks). In contrast, our \cvds induce PG coverings and are 
thus limited to coverings whose mean block size is given by the expression given in~\Cref{mean_variance_thm}, over $v'$ and $k'$ 
such that there is a PG covering for $(v',k',t)$.

\paragraph{Challenge: memory consumption}
The main challenge in computing \cvds is that it requires to compute a PG covering for large values of $v'$ and $k'$, which poses a high memory overhead. To illustrate, in our experiments, \tool uses a \cvd induced from a PG covering for $(v'=1508598, k'=88741, t=5)$. If \tool stored this covering in the memory, it would require $124$GB of memory, assuming each number in a block takes a byte. To cope, \tool computes the partially-induced covering during the PG covering construction. 
However, even the partially-induced coverings can consume a lot of memory, since the number of blocks can be large. Calzone faced a similar challenge and coped by restricting the size of the covering designs to at most $10^7$, which allowed it to keep all coverings in the
covering database. While \tool could take a similar approach, this would prevent it from picking \cvds of larger size which overall may lead to a lower analysis time (since they will require fewer refinements). Instead, \tool generates a \cvd \emph{on-the-fly} and uses the covering database only for the refinements, which tend to require coverings of significantly smaller size than the first covering.  
Another advantage of 
building the \cvd on-the-fly is that it enables \tool 
to analyze any classifier over any image dimension, without any special adaptation. This is in contrast to Calzone, which 
requires to extend its covering database upon every new image dimension $v$.

\section{CoVerD}\label{sec:CoVerDCalzone}
In this section, we present \tool, our $L_0$ robustness verifier. 
We first describe our system and its components and then provide a running example.

 \begin{figure*}[t]
    \centering
  \includegraphics[width=1\linewidth, trim=0 105 0 0, clip, page=1]{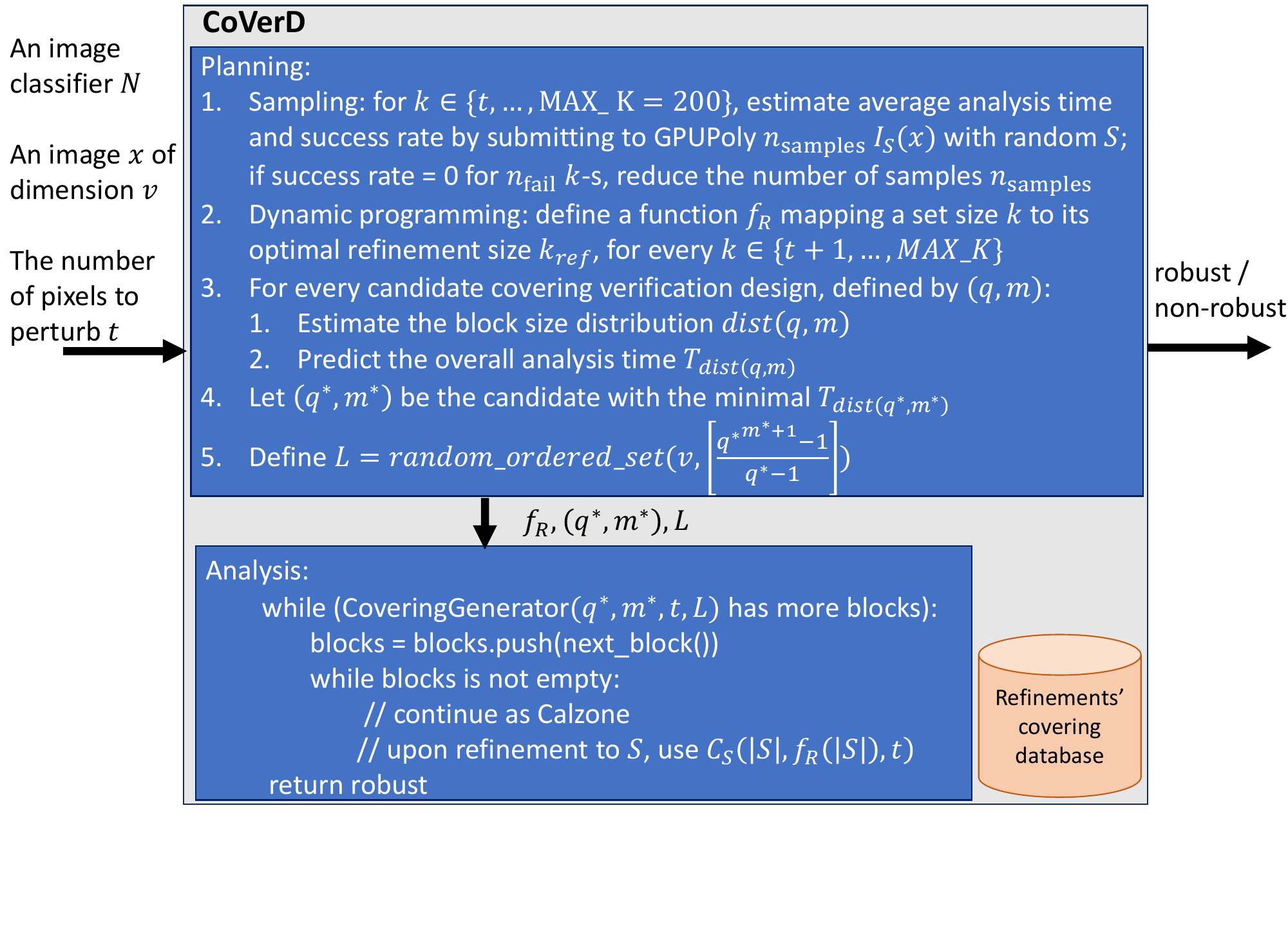}
    \caption{\tool: An $L_0$ robustness verifier.}
    \label{fig:coverd}
\end{figure*}

\subsection{Our System}
\Cref{fig:coverd} shows \tool that, given an image classifier $N$, an image $x$ with $v$ pixels, and the maximal number of perturbed
 pixels $t$, returns whether $N$ is robust in the $t$-ball of $x$.
We next describe its planning and analysis components.

\paragraph{Planning} The planning component consists of several steps. First, it samples sets of different sizes $k$ to estimate the success rate and average analysis time of their respective neighborhoods, like Calzone. Since \tool considers \cvds, it can observe larger block sizes than Calzone, thus the maximal sampled set size is $\text{MAX\_K}=200$, unlike $99$ in Calzone. Because of the larger bound, \tool is likely to observe many more $k$ values whose success rate is zero. To save execution time while still enabling to determine the success rate and average analysis time of large $k$ values, \tool reduces the number of samples 
after observing $n_\text{fail}$ times $k$ values whose success rate is zero.
Second, the planning component relies on Calzone's dynamic programming for computing a K strategy, but uses it differently. 
Since \tool begins the analysis from a \cvd consisting of different sized blocks, there is no single K strategy. Instead, it runs Calzone's dynamic programming for every $k \in \{t+1,\dots ,\text{MAX\_K}\}$ to define a function $f_{R}$ mapping every set size $k$ to the best set size to use upon a refinement of a set of size $k$. 
Then, the planning component iterates over every candidate \cvd and picks the best \cvd for the analysis. 
It picks between the candidates \emph{without} constructing them, as the construction is time and memory intensive and we wish to execute it only for the chosen candidate. 
To pick the best candidate, it leverages two observations. First, a \cvd candidate is uniquely defined by the parameters of the PG covering, $(q,m)$ (formally, its parameters are $(q,m,t)$ but $t$ is identical in all our PG coverings), so it suffices to pick a pair $(q,m)$ which can later be used to construct the \cvd.
Second, to predict the \cvd with the minimal analysis time, only the \emph{block sizes} are needed.
In~\Cref{sec:ExtendingDP}, we describe how to estimate a \cvd's block size distribution $dist(q,m)$ and estimate its analysis time $T_{dist(q,m)}$, in order to predict the best \cvd.
Given the best candidate $(q^*,m^*)$, it randomly samples an ordered set $L$ of $v$ indices from $v'$, which is a function of $(q^*,m^*)$.

\paragraph{Analysis} After determining the best $(q^*,m^*)$, $L$, and the refinement mapping $f_R$, \tool continues to analyze the robustness of the classifier $N$ in the $t$-ball of the given image $x$. The analysis constructs the \cvd on-the-fly block-by-block. Technically, there is a covering generator that constructs the blocks one-by-one. Every block is pushed to the stack of blocks to verify, and then the analysis proceeds as Calzone. That is, the block is popped, submitted to GPUPoly, and if needed, refinement is executed. After the block is verified (directly or by refinement), the next block in the \cvd is obtained from the covering generator. We note that although \tool could use \cvds for refinements, 
the coverings for refinements are smaller than the first covering since these coverings are for triples $(\tilde{v},\tilde{k},t)$ where $\tilde{v}$ is typically few dozens and at most $\text{MAX\_K}=200$, whereas the first covering is for a triple $(\tilde{v},\tilde{k},t)$ where $\tilde{v}$ is the image dimension.
Like Calzone, \tool parallelizes the analysis on GPUs. Thus, our covering generator generates disjoint parts of the covering, described in~\Cref{sec:CoveringConstructor}.

\subsection{Choosing a Covering Verification Design}\label{sec:CoveringDistributionEstimator}
In this section, we describe how \tool predicts the \cvd that enables \tool to minimize the overall analysis time. 
We begin with describing the \cvd candidates, then describe how \tool estimates their block size distributions, and finally explain how \tool predicts the \cvd leading to the minimal analysis time.

\paragraph{Candidates} 
A \cvd candidate is defined by the PG covering from which it is partially-induced.
Recall that a PG covering is defined for triples $(v',k',t)$, where $v'=\frac{q^{m+1}-1}{q-1}$ and $k'=\frac{q^{t}-1}{q-1}$ for a 
prime power $q$ and $m\geq t\geq 2$.
By~\Cref{mean_variance_thm}, given a PG covering, the mean block size of the \cvd has a closed-form expression $\mu_{v',k',v} = \frac{vk'}{v'} = \frac{v(q^{t}-1)}{q^{m+1}-1}$.
By this expression, given $q$, as $m$ increases $\mu_{v',k',v}$ decreases, and given $m$, as $q$ increases $\mu_{v',k',v}$ decreases. 
Further, this expression approaches $0$ for high values of $q$ or $m$. 
Thus, to obtain a \emph{finite} set of candidates, we provide a positive lower bound on $\mu_{v',k',v}$, denoted $\text{MIN\_K}$ (our implementation sets it to $t$). 
That is,  
the finite set of candidates \tool considers is:
$$\{(q,m)\in \mathbb{N}^2 \mid q \text{ is a prime power},\ m\geq t,\ v'\geq v,\ \mu_{v',k',v}\geq \text{MIN\_K} \}$$

\begin{figure*}[t]
\centering 
\begin{subfigure}[t]{.49\textwidth}
 \centering
\includegraphics[width=0.955\linewidth]{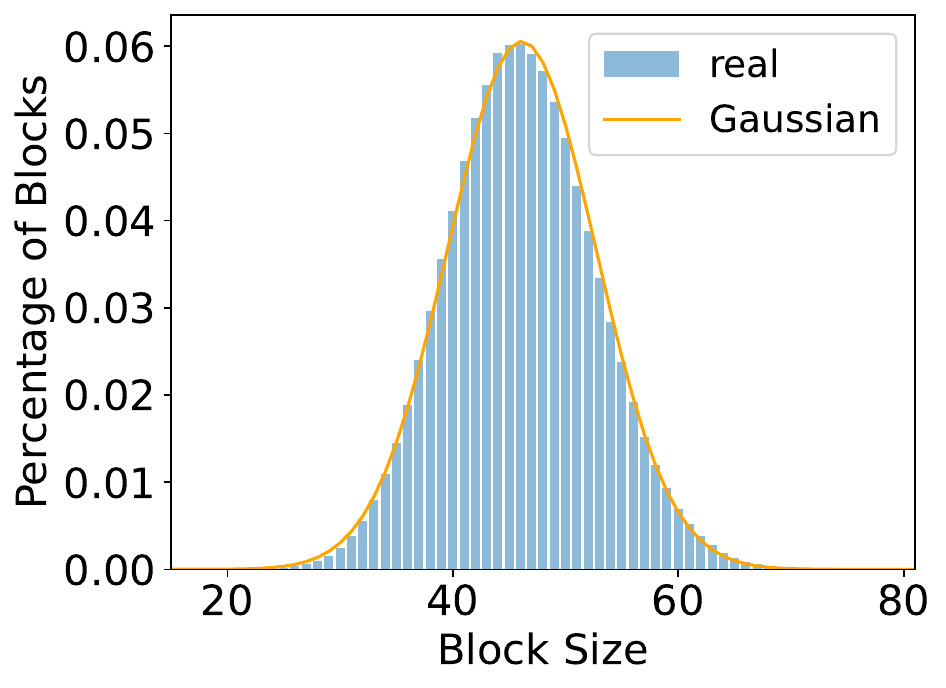}
  \caption{The block size distribution of a \cvd and the respective Gaussian distribution.}
  \label{fig:real_hist_normal_pdf}
\end{subfigure}
\begin{subfigure}[t]{.49\textwidth}
 \centering
\includegraphics[width=1\linewidth]{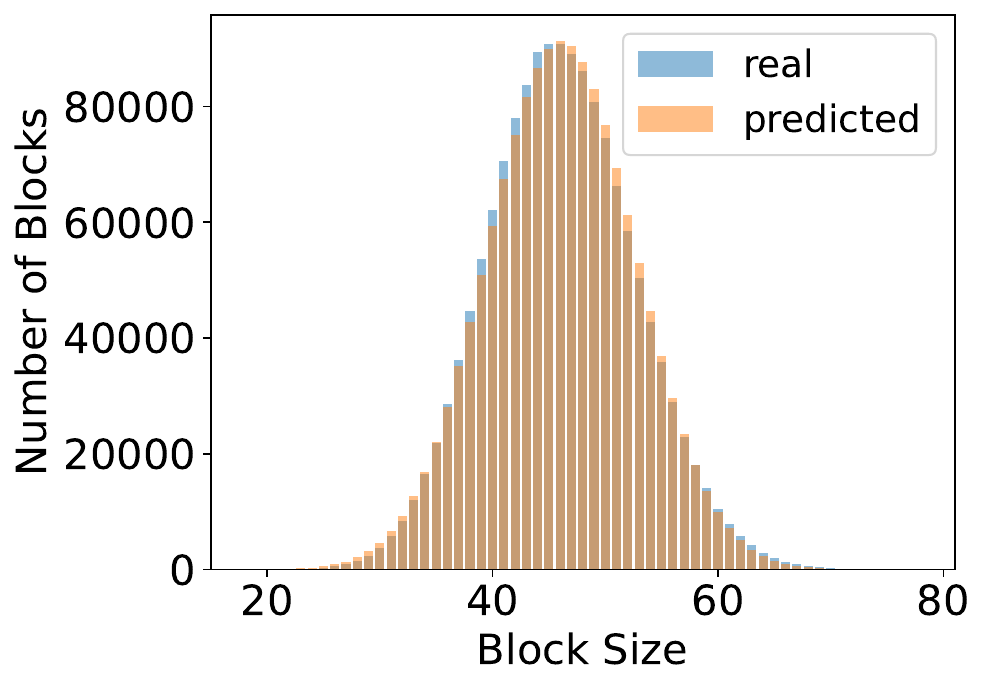}
  \caption{Our estimated block size distribution vs. the distribution of a respective \cvd.}
  \label{fig:real_vs_predicted_hist}
\end{subfigure}
\caption{Block size distributions.}
\end{figure*}

\paragraph{Estimating the block size distribution} For every \cvd candidate, defined by $(q,m)$, \tool estimates the distribution of its block sizes. While \Cref{mean_variance_thm} provides expressions for the mean block size and the variance, it does not define the block size distribution.
We empirically observe that our \cvds have the property that the distribution of their block sizes resembles a discrete approximation of a Gaussian distribution with mean $\mu_{v',k',v}$ and variance $\sigma_{v',k',v}^2$. The higher the mean and the number of blocks, the higher the resemblance.
\Cref{fig:real_hist_normal_pdf} visualizes this resemblance for a \cvd, with $v=784$, induced from a PG with parameters $q=17$, $m=5$, and $t=5$.
We believe this resemblance exists because a \cvd is partially-induced from a PG covering given \emph{a random set of numbers $L$}.
This resemblance may not hold for other choices of $L$, for example for the choice of $L$ proposed by~\cite{Ray-Chaudhuri1968}, which compute a partially-induced covering whose maximal block size is bounded (unlike our \cvd). Because of this resemblance, we model the block size as drawn from the Gaussian distribution with the true mean and variance $\mathcal{G}\left(\mu_{v',k',v},\sigma_{v',k',v}^2\right)$.
Even if this modeling is imprecise, in practice, it is sufficient to allow \tool identify the candidate \cvd leading to the minimal analysis time.
Formally, given a \cvd candidate defined by $(q,m)$, 
the distribution of the block sizes is $dist(q,m)=\{N^{q,m}_k\mid  k \leq \text{MAX\_K}\}$, where $N^{q,m}_k$ is our estimation of the number of blocks of size $k$ in this \cvd.
We define the probability that a block size in this \cvd is of size $k$ as: 
 $\mathbb{P}(k-0.5< Z\leq k + 0.5)=\Phi\left(\frac{(k+0.5)-\mu_{v',k',v}}{\sigma_{v',k',v}}\right) - \Phi\left(\frac{(k-0.5)-\mu_{v',k',v}}{\sigma_{v',k',v}}\right)$, where $Z\sim \mathcal{G}\left(\mu_{v',k',v},\sigma_{v',k',v}^2\right)$ and $\Phi$ is the cumulative distribution function (CDF) of a Gaussian distribution with mean $0$ and variance $1$.
  The number of blocks $b^{q,m}$ is identical to the number of blocks in the PG covering, which has a closed-form expression~\cite{Gordon95newconstructions}.  
Thus, the estimated number of blocks of size $k$ is: 
$\Tilde{N}^{q,m}_k = b^{q,m}\cdot \mathbb{P}(k-0.5< Z \leq k+0.5)$.
To make the estimated number an integer, we define $N^{q,m}_k$ as the floor of $\Tilde{N}^{q,m}_k$ and add $1$ with probability of the remainder:
$N^{q,m}_k = \left\lfloor \Tilde{N}^{q,m}_k \right\rfloor + X$ where $X \sim  \text{Bern} \left(\Tilde{N}^{q,m}_k - \left\lfloor \Tilde{N}^{q,m}_k \right\rfloor\right)$.
 \Cref{fig:real_vs_predicted_hist} visualizes how close our estimation of the block size distribution is to the distribution of the \cvd shown in \Cref{fig:real_hist_normal_pdf}.
We note that \tool considers a candidate and estimates its block size distribution only if its 
estimated number of overly large blocks (larger than $\text{MAX\_K}$) is close to zero. Formally, it considers candidates that satisfy 
$b^{q,m}\cdot \left ( 1- \Phi\left(\frac{\text{MAX\_K}-\mu_{v',k',v}}{\sigma_{v',k',v}}\right) \right ) \leq \epsilon$, where $\epsilon$ is a small number. This is the reason that $dist(q,m)=\{N^{q,m}_k\mid  k \leq \text{MAX\_K}\}$ consists of estimations only for blocks whose size is at most $\text{MAX\_K}$.

\paragraph{Predicting the best \cvd}\label{sec:ExtendingDP}
Given the candidates and their estimated block size distributions, \tool chooses the \cvd which will enable \tool to minimize the overall analysis time. To this end, it predicts for every candidate \cvd its overall analysis time. The prediction 
relies on: (1)~the estimated number of blocks $N^{q,m}_k$ of size $k$, (2)~the average analysis time of a block of size $k$, denoted $k\_array[k][time]$  (given by the initial sampling), (3)~the fraction of the non-robust blocks of size $k$, which is one minus the success rate of $k$, denoted $k\_array[k][success]$ (given by the initial sampling) and~(4) the analysis time of refining a non-robust block of size $k$, denoted $T(k)$ (given by the dynamic programming, as defined in~\cite{Shapira23}). Similarly to Calzone's dynamic programming, the analysis time is the sum of the analysis time of verifying all blocks in the \cvd and the analysis time of the refinements of the non-robust blocks:
$$T_{dist(q,m)}=\sum_{k=t}^{\text{MAX\_K}}N^{q,m}_k \cdot \left (k\_array[k][time] + (1-k\_array[k][success])\cdot T(k) \right )$$
This computation ignores blocks of size less than $t$ since they do not cover any subset of size $t$ and need not be analyzed to prove $L_0$ robustness. After predicting the analysis time of every candidate,  
\tool picks the candidate with the minimal time. 
\subsection{Constructing a Covering Verification Design}\label{sec:CoveringConstructor}
In this section, we present our covering generator that computes a \cvd. 
The covering generator operates as an independent process, one for every GPU, that outputs blocks a-synchronically. 
At every iteration, every GPU worker obtains a block from its covering generator, analyzes it with GPUPoly, and refines if needed. If the block is robust or its refinement does not detect an adversarial example, the GPU worker obtains the next block from the covering generator. 
The covering generator relies on the chosen \cvd's parameters $q$ and $m$ and the ordered set $L$ from the planning component.
It computes the PG covering for ($q$, $m$, $t$) block-by-block and induces it to obtain a \cvd.
Generally, its construction follows the meta-algorithm of generating PG coverings described in~\cite{Gordon95newconstructions}. The novel parts are our implementation of inducing blocks immediately upon generating them and partitioning them to enable their analysis to proceed in parallel over the available GPUs.
We next describe the covering generator.

\begin{algorithm}[t]
\DontPrintSemicolon
\KwIn{PG parameters ($q$, $m$, $t$), an ordered set of $v$ indices $L=[s_1,\dots,s_v]$ which is a subset of $[\frac{q^{m+1}-1}{q-1}]$, and an index of a GPU $i_{GPU}$.}
\KwOut{A stream of the covering verification design's blocks.}

$\forall i\in [v].\ P[:,i]$ = a unique vector in $\mathbb{F}_q^{m+1}$ computed for $s_i$\tcp*{$P \in \mathbb{F}_q^{(m+1)\times v}$}\label{conp}
$\mathcal{M}=[M \in  \mathbb{F}_q^{(m-t+1)\times (m+1)} \mid \text{$M$ is full rank and in reduced row echelon form}]$\;\label{conmat}
\For{$j=0$; $j < |\mathcal{M}|$; $j++$}{\label{conbegin}
    \lIf{$j \text{ modulo } GPUs \neq i_{GPU}$}{continue}\label{conskip}
    $R$ = $\mathcal{M}[j]\times P$\;\label{conmul}
    block = $\left\{i \in [v] \mid R[:,i] = \vec{0}\right\}$\tcp*{generate induced block}\label{coninduce}
    output block
    }\label{conend}
\caption{CoveringGenerator($q$, $m$, $t$, $L$, $i_{GPU}$)}\label{alg:ComputeCovering1}
\end{algorithm}

\Cref{alg:ComputeCovering1} shows the algorithm of our covering generator.
It takes as input the PG parameters $(q, m, t)$, an ordered set $L$ of $v$ indices from $[v']$ (where $v'=\frac{q^{m+1}-1}{q-1}$), and the GPU index $i_{GPU}$.
As described in~\Cref{sec:approach}, a PG construction for ($v'$, $k'$, $t$) views $v'$ as the number of points in the geometry. Formally, given the finite field $\mathbb{F}_q$ of order $q$, we identify the points of the geometry as a subset $W\subset \mathbb{F}_q^{m+1}$ of size $v'$
(technically, these points are representatives of equivalence classes over $\mathbb{F}_q^{m+1}$, as explained in~\cite{Gordon95newconstructions}). 
To later partially-induce the covering using $L$, \Cref{alg:ComputeCovering1} 
maps every index in $L$ to a unique point in $W$ and stores all points (column vectors) in a matrix $P$ (\Cref{conp}).
Then, \Cref{alg:ComputeCovering1} begins to construct the PG covering by computing flats (linear subspaces) of dimension $t-1$, each containing $k'=\frac{q^t-1}{q-1}$ points.
As described in~\cite{Gordon95newconstructions}, every block in the PG covering is a solution (a set of points in $W$) to $m-t+1$ independent linear equations over $m+1$ variables.
Such a linear system can be represented as a full rank matrix, where its solutions are vectors in the matrix's null space.
Thus, to compute the blocks in the PG covering, \Cref{alg:ComputeCovering1} defines a set of matrices $\mathcal{M}$, each is over $\mathbb{F}_q$, of dimension $(m-t+1)\times(m+1)$, and full rank (equal to $m-t+1$).
Each matrix has exactly $k'$ points in $W$ in its null space. These points form a PG block (a flat of dimension $t-1$). 
 To avoid block duplication, the matrices in $\mathcal{M}$ need to have different null spaces. 
Thus, \Cref{alg:ComputeCovering1} considers matrices in reduced row echelon form, i.e.,   
$\mathcal{M}$ 
is all full rank $(m-t+1)\times(m+1)$ matrices over $\mathbb{F}_q$ in reduced row echelon form (\Cref{conmat}). The covering generator then iterates these matrices.
To avoid a high memory overhead,
 the matrix $\mathcal{M}[j]$ is generated only upon reaching its index $j$.
If $j$ belongs to the disjoint part of the given GPU, its induced block is generated (\Cref{conskip}). 
To construct a PG block, one needs to compute all the points in the null space of $\mathcal{M}[j]$.
However, the generator requires only the partially-induced blocks. Thus, it \emph{immediately induces} the block by obtaining all points $s_i$, for $i\in [v]$, whose respective point $P[:,i]$ belongs to the null space of $\mathcal{M}[j]$.
To this end, it defines $R$ as the multiplication of $\mathcal{M}[j]$ and $P$ (\Cref{conmul}), forms the induced block by identifying the points that are in the null space of $\mathcal{M}[j]$ (i.e., every $s_i$ satisfying
 $R[:,i] = \vec{0}$), and makes the induced block a subset of $[v]$ by mapping every $s_i$ in the induced block to $i$ (\Cref{coninduce}).  
\subsection{A Running Example}
In this section, we describe a real execution of \tool, for an MNIST image, a fully-connected network (6$\times$200\_PGD in~\Cref{sec:evaluation}), and $t=4$. 
\tool begins with the planning component. It first estimates the success rate and average analysis time of blocks.
For every $k\in \{4,5,\dots,200\}$, it samples blocks $S$ (subsets of $[784]$) of size $k$ and submits their neighborhood $I_S(x)=\{x' \in [0,1]^v \mid \forall i\notin S.\ x'_i=x_i\}$ to GPUPoly. 
Based on all samples for $k$, it estimates the success rate and the average analysis time. 
For instance, for $k=34$ the success rate is $94.05\%$ and the average analysis time is $16.19$ms, while for $k=41$ they are $65.85\%$ and $16.96$ms. 
Then, \tool runs Calzone's dynamic programming to map every $k \in \{5,6,\dots,200\}$ to the refinement size. For example, $k=34$ is mapped to $28$ and $k=41$ to $33$. Next, \tool determines the \cvd for the first covering, out of $50$ candidates. For each candidate, it predicts the block size distribution and the respective overall analysis time of this candidate. 
To this end, it computes the mean, variance, and number of blocks using the closed-form expressions.
For example, the \cvd of the candidate $(q=23,m=4)$ has mean block size $34.087$, variance $32.518$ and $292,561$ blocks.
The \cvd of $(q=19,m=4)$ has mean block size $41.263$, variance $38.867$, and $137,561$ blocks. Although the second candidate has less than half the number of blocks of the first candidate, \tool predicts that using the first candidate will enable a faster analysis. This is because its success rate is significantly higher and thus it will require fewer refinements (e.g., the success rate of its mean block size is $94.05\%$, whereas the second candidate's success rate of the mean block size is $65.85\%$). The estimated analysis times (in minutes) are $T_{dist(23,4)}=21.20$ and $T_{dist(19,4)}=27.92$.
The last step of the planning component samples an ordered set $L$ of size $784$ (the number of pixels in the MNIST image) from 
$[\frac{23^5-1}{23-1}] = [292561]$. 
In total, the planning component takes 63.5 seconds. 

Then, \tool continues to the analysis component. It starts by creating eight instances of the covering generator (\Cref{alg:ComputeCovering1}), one for each GPU. A covering generator creates blocks for its GPU one-by-one, given $q^*=23$, $m^*=4$, $t=4$ and $L$. For every \cvd block $S$, the GPU worker defines its neighborhood $I_S(x)$ and submits to GPUPoly. 
If GPUPoly verifies successfully, the next \cvd block is obtained.
If GPUPoly fails proving robustness, $S$ is refined. 
As example, if a block $S$ of size $34$ is refined, the analysis pushes to the stack all blocks in the covering $C_S(34,28,4)$, which is the covering for $(34,28,4)$ that is in the covering database, where the numbers are renamed to range over the numbers in $S$. 
In this example, GPUPoly is invoked $659,326$ times, where $44\%$ of these calls are for blocks in the \cvd. The maximal size of block submitted to GPUPoly is $62$ and the minimal size is $8$. In particular, \tool did not submit any block of size $t=4$ (i.e., there are no calls to the MILP verifier).    
The analysis takes 23.49 minutes, which is only $10.8\%$ higher than the estimated time.  
\section{Evaluation}\label{sec:evaluation}
In this section, we describe the experimental evaluation of \tool on multiple datasets and networks and compare it to Calzone.

\paragraph{Implementation and setup}
We implemented \tool\footnote{\url{https://github.com/YuvShap/CoVerD}} as an extension of Calzone\footnote{\url{https://github.com/YuvShap/calzone}}. Experiments ran on a dual AMD EPYC 7713 server, 2TB RAM, eight NVIDIA A100 GPUs and Ubuntu 20.04.1. We evaluate \tool on the networks evaluated by Calzone, whose architectures are described in ERAN\footnote{\url{https://github.com/eth-sri/eran}}.
We consider networks trained for popular image datasets: MNIST and Fashion-MNIST, consisting of $28 \times 28$ greyscale images, and CIFAR-10, consisting of $32\times 32$ colored images. 
\tool's hyper-parameters are: the maximal block size is $\text{MAX\_K} = 200$, the number of samples is initially $n_\text{samples}=400$
and after $n_\text{fail}=10$ failures, it is reduced to $n_\text{samples}=24$, and the bound on the estimated number of overly large blocks is $\epsilon= 0.01$. 
Our covering database, used for the refinement steps, contains coverings for $v,k\leq 200$, $t\leq 6$. The covering sizes are restricted to at most $500,000$ blocks. 
This limitation is stricter than Calzone, which limited to $10^7$, but in practice this is unnoticeable since \tool only uses the coverings for refinements, and even Calzone typically refines to coverings whose size is at most $500,000$. 
Like Calzone, the database consists of coverings computed by extending coverings from the La Jolla Covering Repository Tables\footnote{\url{https://ljcr.dmgordon.org/cover/table.html}} using
construction techniques from~\cite[Section 6.1]{Gordon95newconstructions}.
 Additionally, our database includes finite geometry coverings (for $v,k\leq 200$, $t\leq 6$) and extends coverings using the dynamic programming of~\cite[Section 5]{Gordon95newconstructions}. 
Like Calzone, We ran \tool with eight GPUPoly instances and five MILP instances, except for the CIFAR-10 network where it ran 50 MILP instances. 
 For the matrix multiplication over finite fields (\Cref{alg:ComputeCovering1}), \tool relies on an effective library~\cite{Hostetter_Galois_2020} and considers only prime numbers for $q$ (since matrix multiplication is too slow for prime powers).

 \begin{figure*}[t]
 \centering
\begin{subfigure}{.33\textwidth}
  \centering
  \includegraphics[width=1\linewidth]{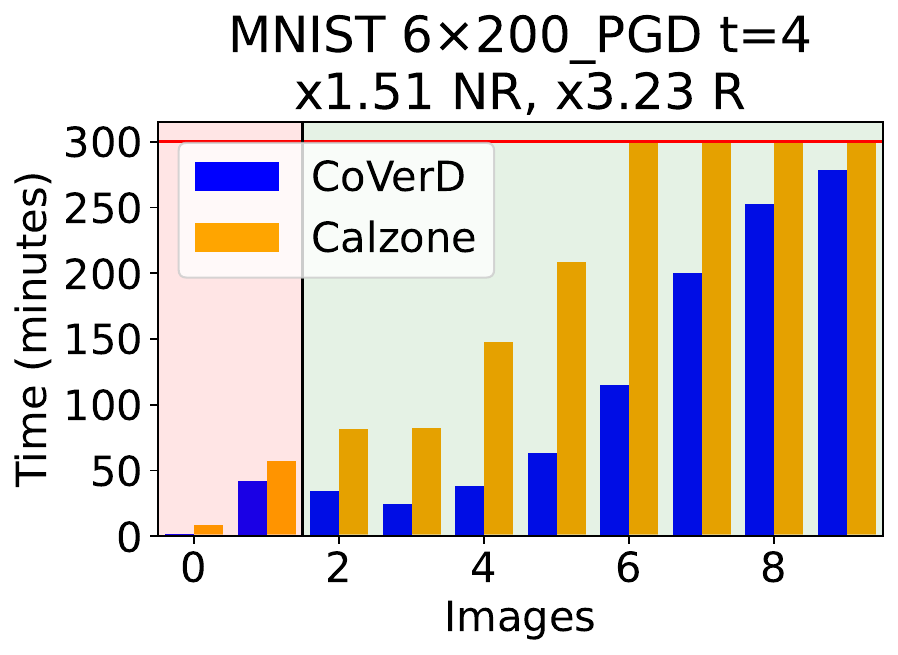}
\end{subfigure}%
\begin{subfigure}{.33\textwidth}
  \centering
  \includegraphics[width=1\linewidth]{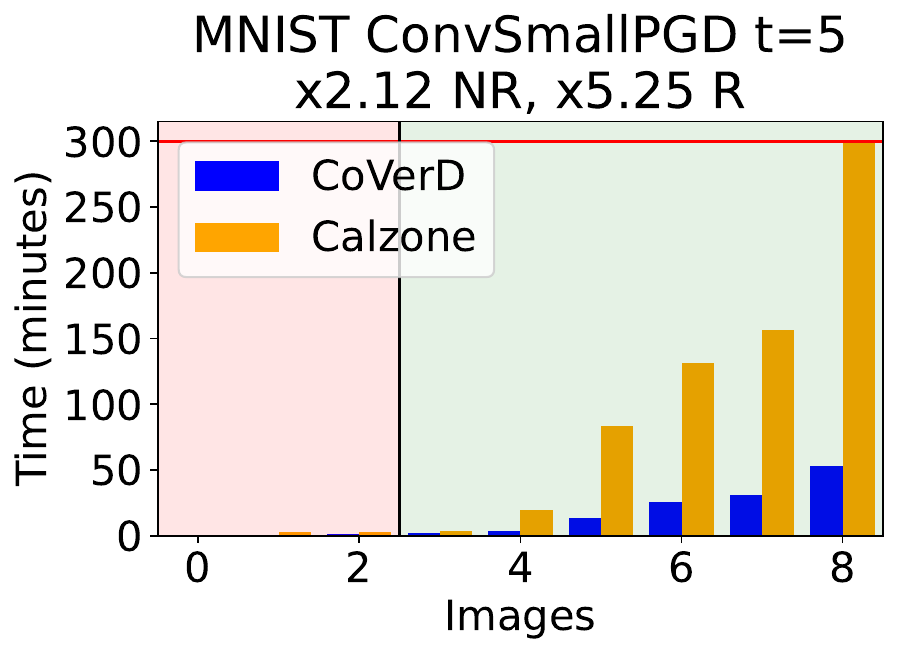}
  \end{subfigure}%
  \begin{subfigure}{.33\textwidth}
  \centering
  \includegraphics[width=1\linewidth]{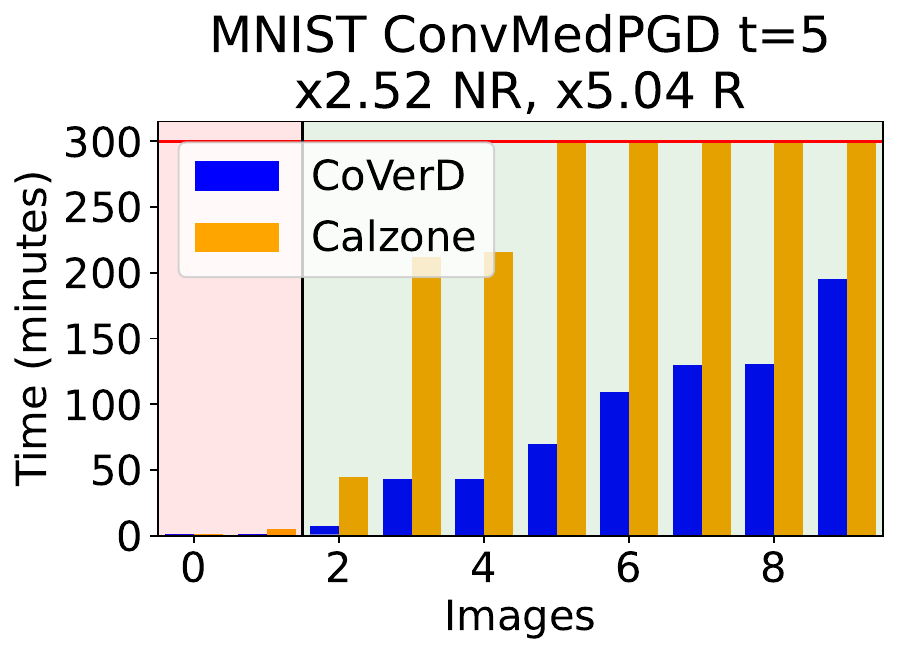}
  \end{subfigure}
\begin{subfigure}{.33\textwidth}
  \centering
  \includegraphics[width=1\linewidth]{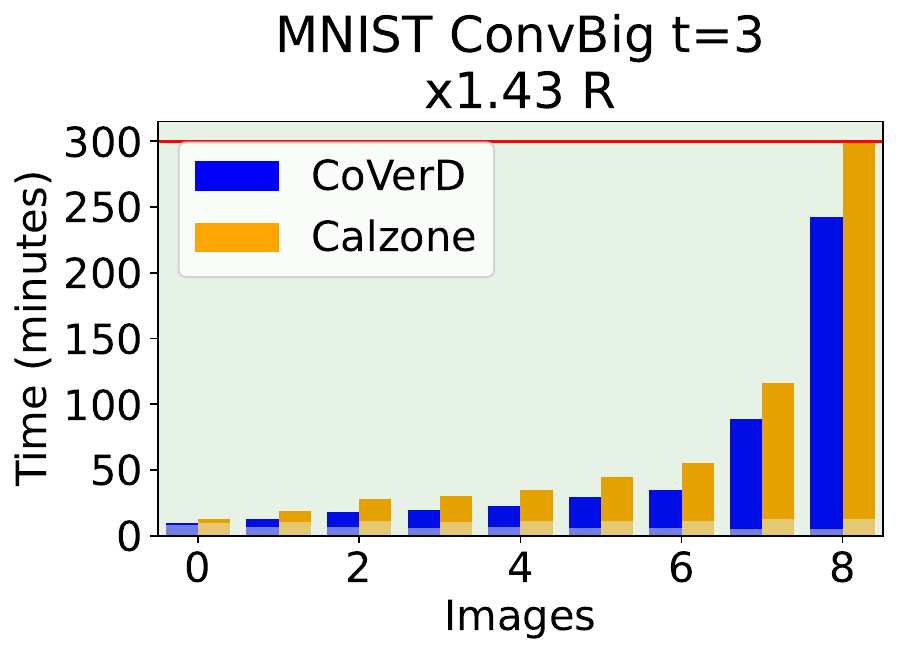}
\end{subfigure}%
\begin{subfigure}{.33\textwidth}
  \centering
  \includegraphics[width=1\linewidth]{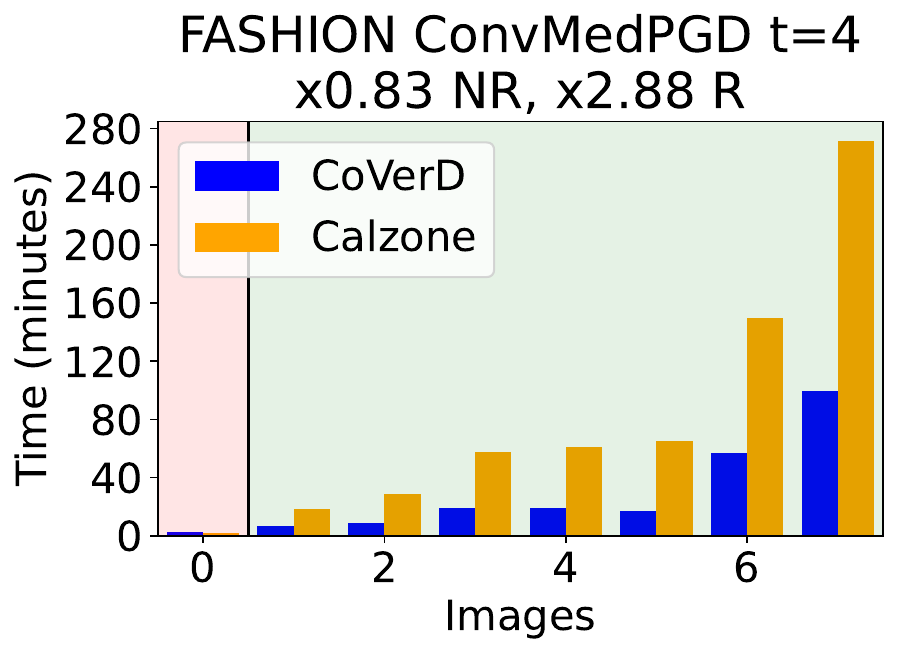}
  \end{subfigure}%
  \begin{subfigure}{.33\textwidth}
  \centering
  \includegraphics[width=1\linewidth]{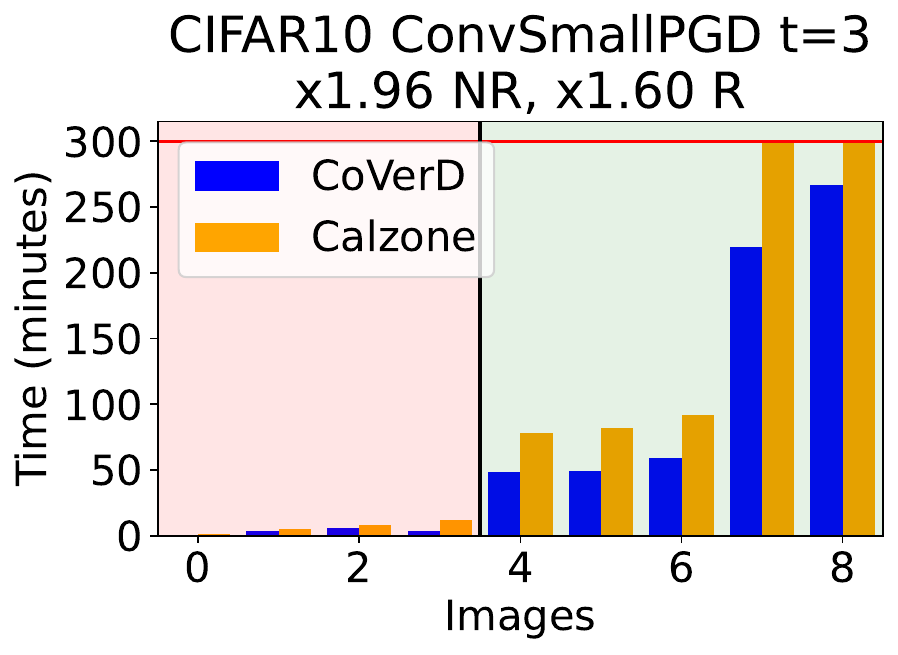}
  \end{subfigure}
  \caption{\tool vs. Calzone on Calzone's most challenging benchmarks.}\label{fig:reschal}
\end{figure*}

\begin{figure*}[t!]
 \centering
\begin{subfigure}{1\textwidth}
  \centering
  \includegraphics[height=3cm,width=1\linewidth]{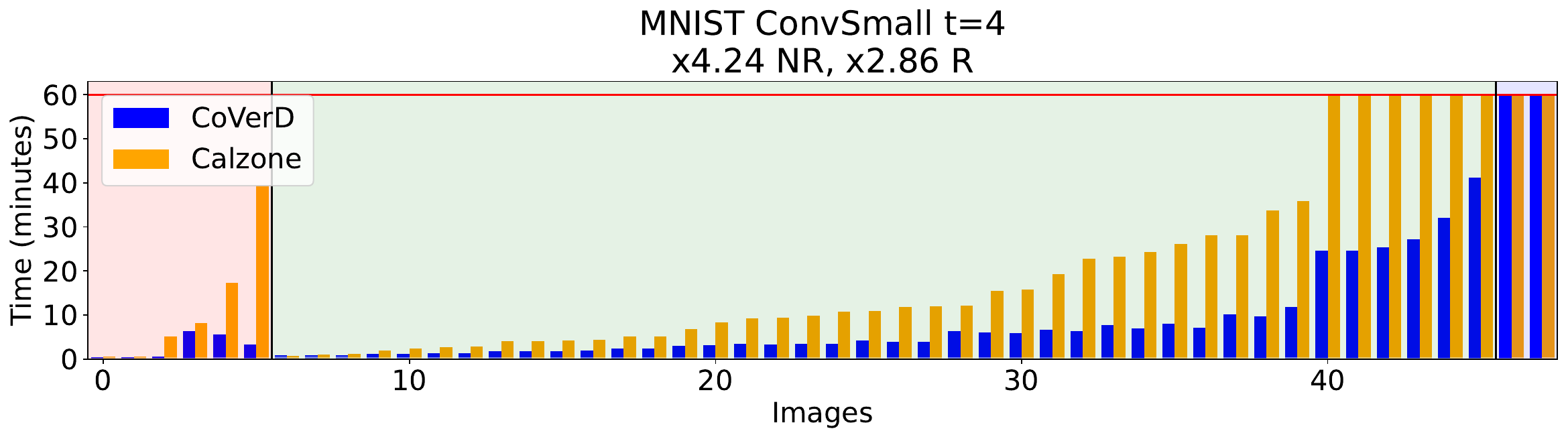}
  \end{subfigure}
\begin{subfigure}{1\textwidth}
  \centering
  \includegraphics[height=3cm,width=1\linewidth]{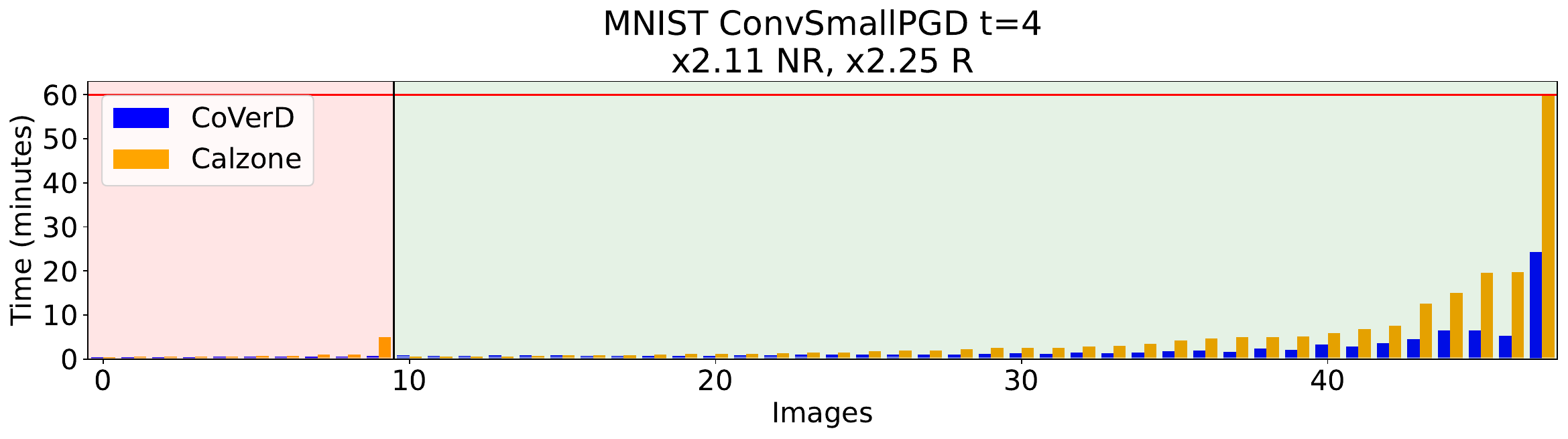}
  \end{subfigure}
  \begin{subfigure}{1\textwidth}
  \centering
  \includegraphics[height=3cm,width=1\linewidth]{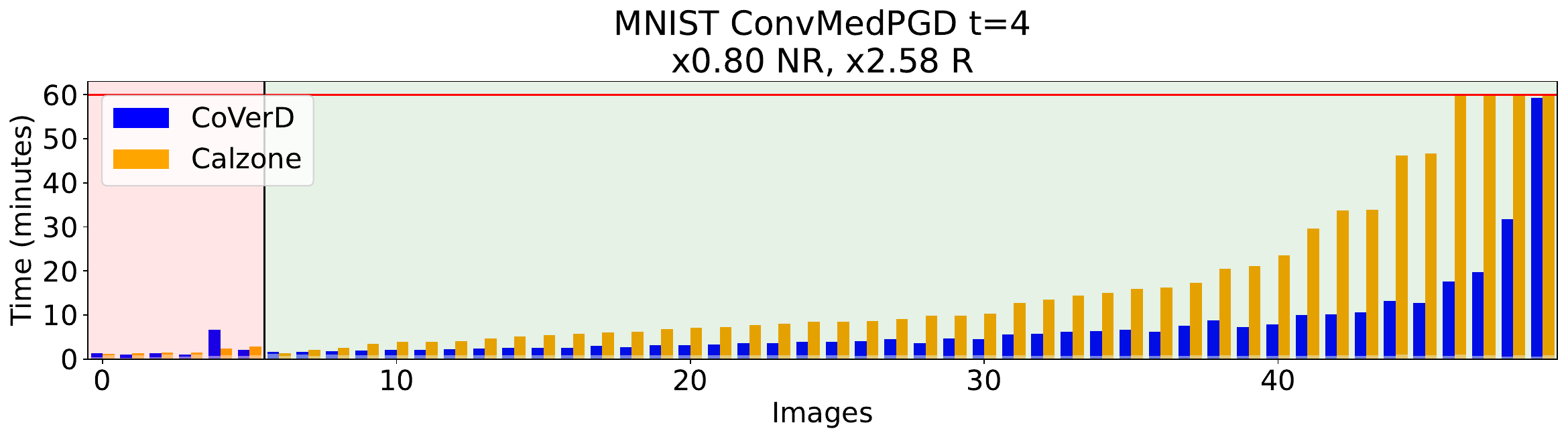}
  \end{subfigure}
  \begin{subfigure}{1\textwidth}
  \centering
  \includegraphics[height=3cm,width=1\linewidth]{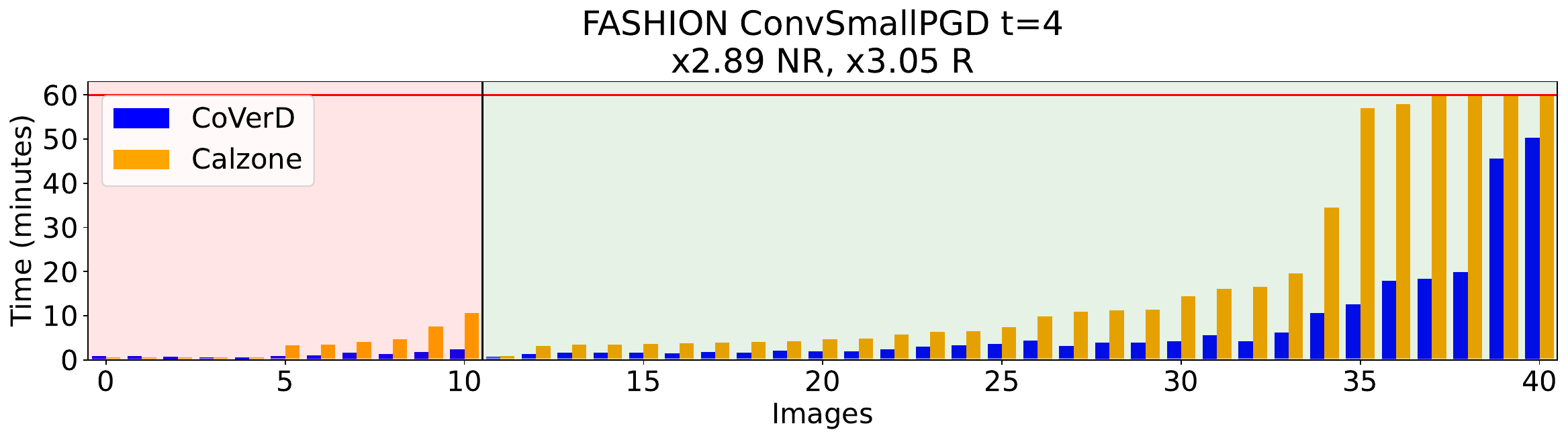}
  \end{subfigure}
  \caption{\tool vs. Calzone for $t=4$.}\label{fig:tfour}
\end{figure*}

\paragraph{Comparison to Calzone}
We begin by evaluating \tool on Calzone's benchmarks (i.e., the same networks, images and timeouts) for $t\geq 3$.
\Cref{fig:reschal} shows the comparison for the most challenging benchmarks of Calzone, and \Cref{fig:tfour} shows comparisons for $t=4$ (the plots for $t=3$ are shown in~\ifthenelse{\boolean{shortver}}{\cite[Appendix A]{our_arxiv}}{\Cref{sec:addres}}). 
For a given network and $t$, the plot shows the execution time in minutes 
of \tool and Calzone for every $t$-ball. The $x$-axis orders the $t$-balls by \tool's output: non-robust (in light red background), robust (in light green background), and timeout (in light blue background, e.g., \Cref{fig:tfour}, top).
 Within each section, the $t$-balls are sorted by their execution time for clearer visuality.
Timeouts, of \tool or Calzone, are shown by bars reaching the red horizontal line. 
The lower part of each bar shows in a lighter color the execution time of the initial sampling (unless it is too short to be visible in the plot). The sampling time is highlighted since Calzone and \tool sample slightly differently: Calzone samples $400$ sets of size $k$, for every $k\leq 99$, while \tool samples up to $k\leq 200$ and reduces the number of samples after observing ten $k$ values whose average success rate is zero. We note that the other computations of the planning component take a few seconds.
The plots' titles include the speedup in the average analysis time of \tool over Calzone for non-robust $t$-balls (NR) and for robust $t$-balls (R).

The plots show that, on the most challenging benchmarks (\Cref{fig:reschal}), 
 \tool is always faster than Calzone, except for two non-robust $t$-balls which \tool completes their analysis within 140 seconds.
  In the plots of~\Cref{fig:tfour}, \tool is always faster than Calzone except for thirteen $4$-balls whose analysis terminates within seven minutes by both verifiers. 
In the other plots (\ifthenelse{\boolean{shortver}}{Figure 9 in~\cite[Appendix A]{our_arxiv}}{\Cref{fig:tthre} in~\Cref{sec:addres}}), where $t=3$, Calzone is sometimes faster, but in these cases the analysis time is typically short. 
In other words, the significance of \tool is in shortening the analysis time of $t$-balls with long analysis time.
On average, \tool is faster than Calzone in verifying robust $t$-balls by 1.3x for $t=3$, by 2.8x for $t=4$, and by 5.1x for $t=5$.

\begin{figure*}[t]
\centering 
\begin{subfigure}[t]{.33\textwidth}
 \centering
\includegraphics[width=1\linewidth]{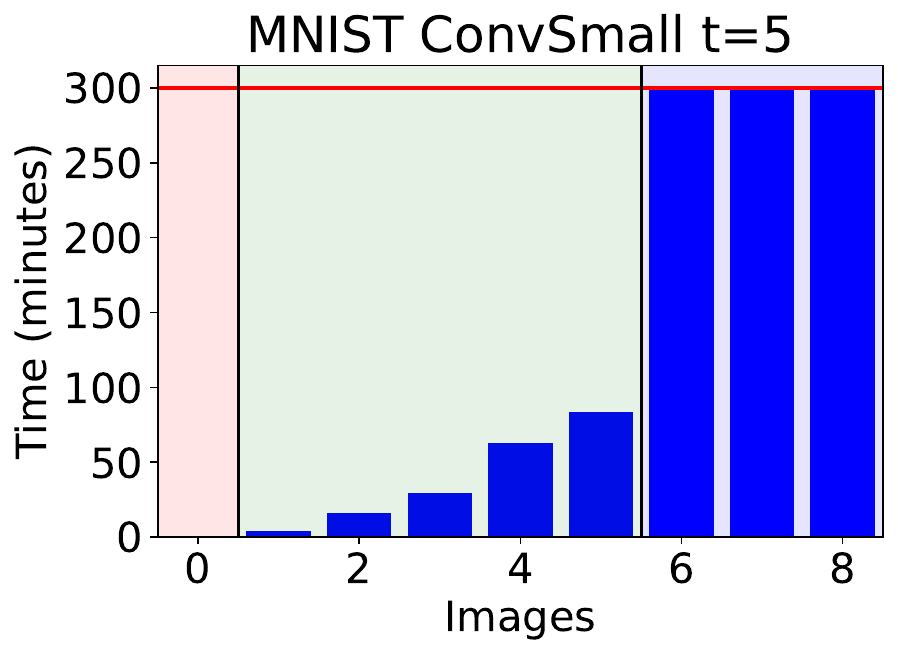}
\end{subfigure}%
\begin{subfigure}[t]{.33\textwidth}
 \centering
\includegraphics[width=1\linewidth]{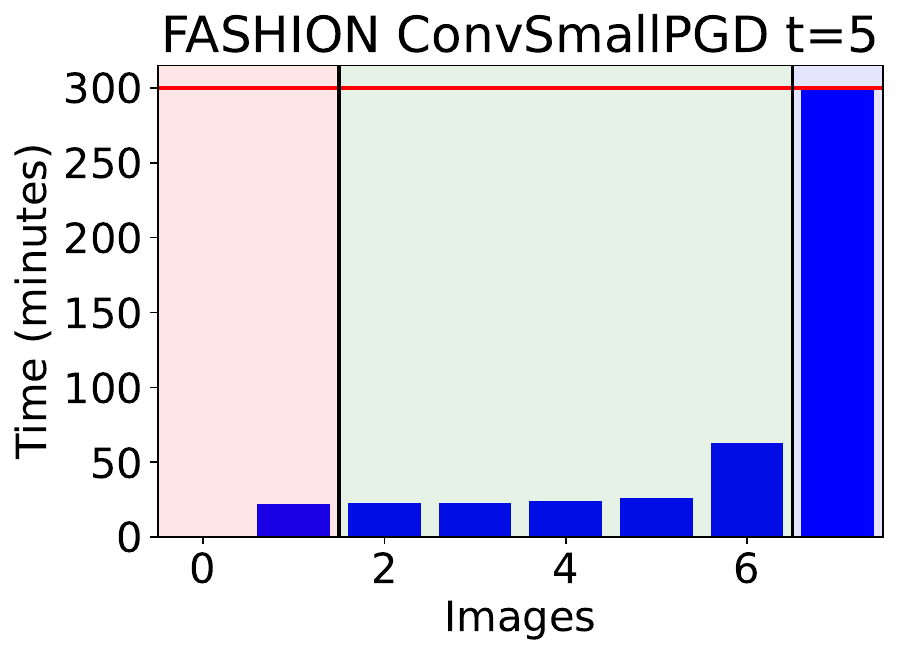}
\end{subfigure}%
\begin{subfigure}[t]{.33\textwidth}
 \centering
\includegraphics[width=1\linewidth]{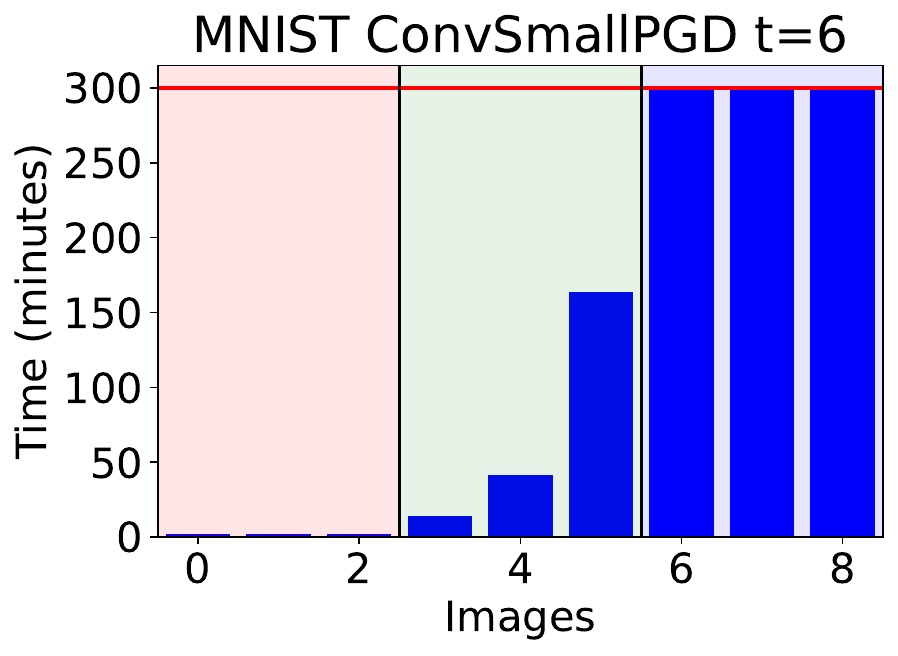}
\end{subfigure}%
\caption{\tool's new benchmarks.}\label{fig:resnew}
\end{figure*}

\paragraph{Challenging benchmarks} 
Next, we show more challenging benchmarks than Calzone. 
We evaluate the robustness of three networks for $t$-balls with larger values of $t$ than Calzone considers, for $t=5$ and for $t=6$ (we remind that Calzone is evaluated for $t\leq 5$). 
Similarly to Calzone's most challenging benchmarks, these benchmarks evaluate \tool for ten images (misclassified images are discarded) and a five hour timeout.
\Cref{fig:resnew} shows \tool's analysis time. 
\tool completes the analysis for $73\%$ $t$-balls. Further, it verifies robustness in some $6$-balls within $42$ minutes. As before, \tool is significantly
faster for non-robust $t$-balls.

We provide additional statistics on \tool in~\ifthenelse{\boolean{shortver}}{\cite[Appendix A]{our_arxiv}}{\Cref{sec:addres}}.
\section{Related Work}
In this section, we discuss the closest related work.

\paragraph{Robustness verification of neural networks}
Many works propose robustness verifiers for neural networks. 
Most works focus on local robustness in $L_\infty$ neighborhoods, defined by a series of intervals~\cite{MullerS0PV21,Tjeng19,SinghGPV19,GehrMDTCV18,KatzBDJK17,AndersonPDC19,WuOZJIGFKPB20,WangZXLJHK21,Ehlers17}. Some verifiers provide
a complete analysis, i.e., they determine whether a network is robust in the given neighborhood~\cite{Tjeng19,KatzBDJK17,Ehlers17}. These approaches typically rely on constraint solving (SAT/SMT solvers or MILP solvers) and thus they often do not scale to large networks. Incomplete verifiers scale the analysis by over-approximating the non-linear computations of the network (e.g., the activation functions) by linear approximations or abstract domains~\cite{MullerS0PV21,SinghGPV19,GehrMDTCV18,WangZXLJHK21}. 
Several local robustness verifiers address $L_2$-balls, e.g., by computing a bound on the network's global or local Lipschitz constant~\cite{Leino21,Huang21}, or $L_1$-balls~\cite{WuZ21,ZhangWCHD18}. Other approaches analyze robustness in $L_p$-balls for $p\in \{0,1,2,\infty\}$ using randomized smoothing~\cite{CohenRK19,YangDHSR020,LiCWC19,SalmanLRZZBY19,DvijothamHBKQGX20}, providing probabilistic guarantees. To the best of our knowledge, Calzone~\cite{Shapira23} is the first work to deterministically verify local robustness in $L_0$-balls. Other works prove robustness in neighborhoods defined by high-level features~\cite{KabahaD22,MohapatraWC0D20,BalunovicBSGV19}. 

\paragraph{Covering and combinatorial designs} \cvd is related to several combinatorial designs: the combinatorial design defined by~\cite{Ray-Chaudhuri1968}, covering designs~\cite{Gordon95newconstructions} and balanced incomplete block designs~\cite{alma990022986490203971}.
Covering designs, in particular finite geometry coverings, have been leveraged in various domains, including file information retrieval~\cite{Ray-Chaudhuri1968}, file organization~\cite{Abraham1968} and coding theory~\cite{Chan1981}. 
General combinatorial designs have also been leveraged in various domains in computer science~\cite{Colbourn1989}.  
\section{Conclusion}
We present \tool, an $L_0$ robustness verifier for neural networks. \tool boosts the performance of a previous $L_0$ robustness verifier by employing several ideas. First, it relies on a covering verification design (\cvd), a new combinatorial design partially inducing a projective geometry covering. Second, it chooses between candidate \cvds without constructing them but only predicting their block size distribution. Third, it constructs the chosen \cvd on-the-fly to keep the memory overhead minimal. We evaluate \tool on fully-connected and convolutional networks. We show that it boosts the performance of proving a network's robustness to at most $t$ perturbed pixels on average by 2.8x, for $t=4$, and by 5.1x, for $t=5$. For $t=6$, \tool sometimes proves robustness within $42$ minutes. 
\bibliographystyle{splncs04.bst}
\bibliography{mybibliography}
\ifthenelse{\boolean{shortver}}{}{\appendix \section{Additional Results}\label{sec:addres}
In this section, we provide additional experimental results.

\begin{figure*}[t!]
 \centering
\begin{subfigure}{0.95\textwidth}
  \centering
  \includegraphics[width=1\linewidth]{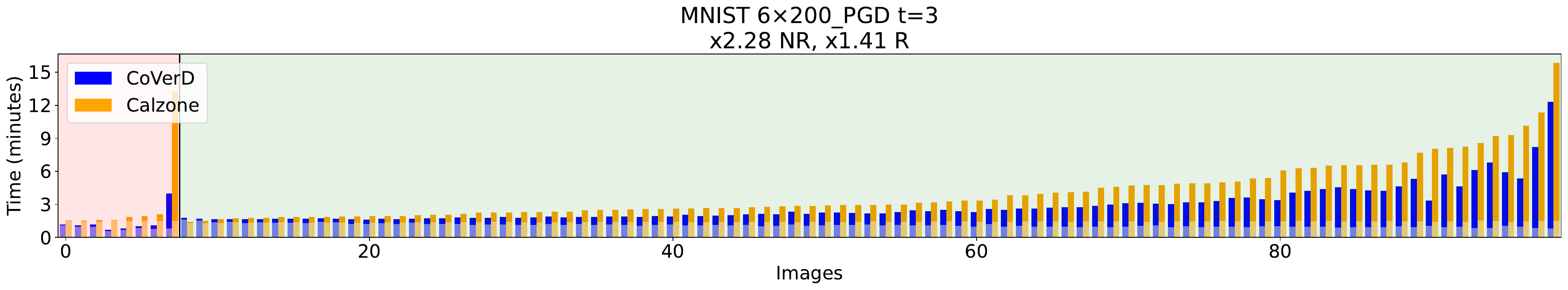}
  \end{subfigure}
\begin{subfigure}{0.95\textwidth}
  \centering
  \includegraphics[width=1\linewidth]{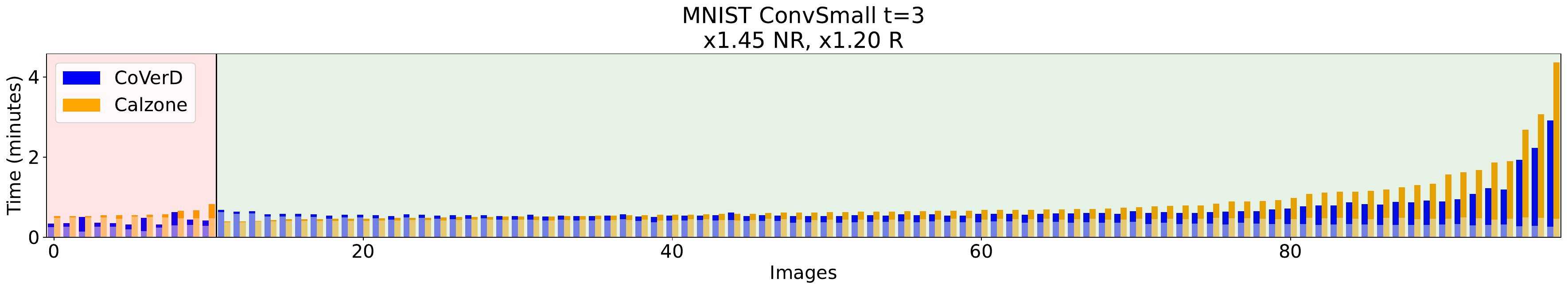}
  \end{subfigure}
  \begin{subfigure}{0.95\textwidth}
  \centering
  \includegraphics[width=1\linewidth]{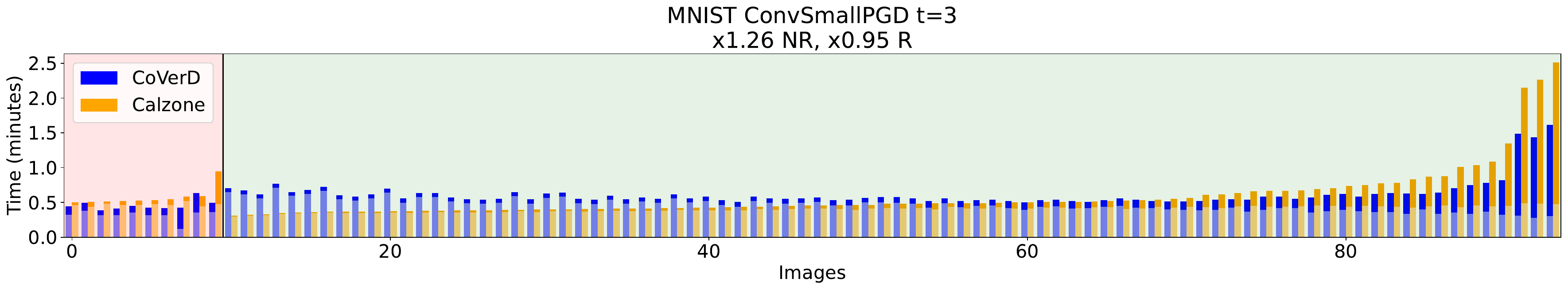}
  \end{subfigure}
  \begin{subfigure}{0.95\textwidth}
  \centering
  \includegraphics[width=1\linewidth]{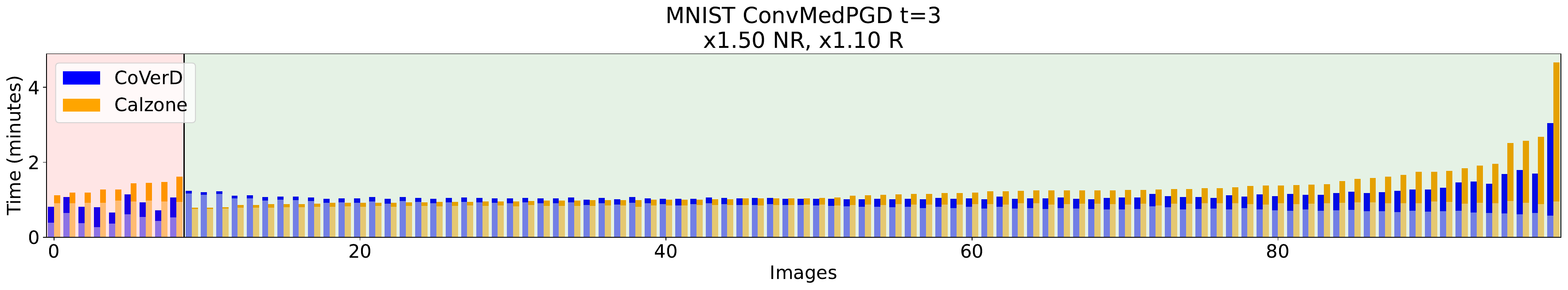}
  \end{subfigure}
  \begin{subfigure}{0.95\textwidth}
  \centering
  \includegraphics[width=1\linewidth]{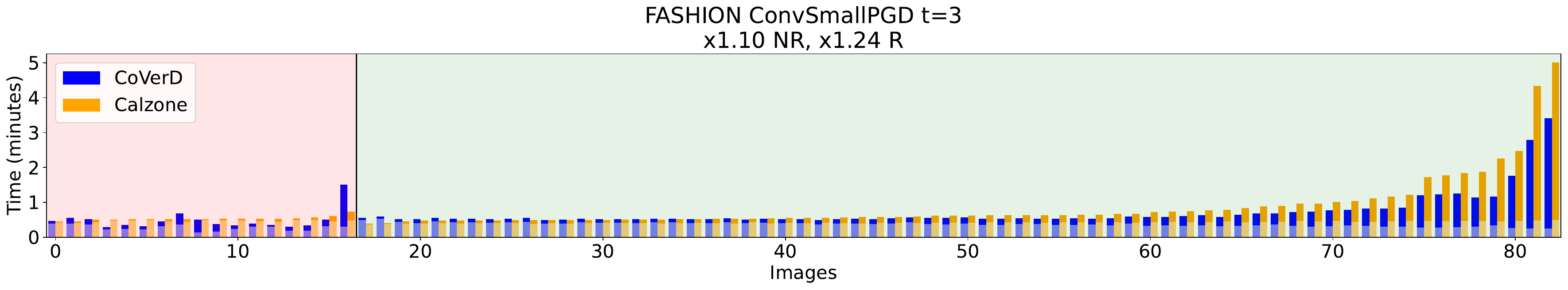}
  \end{subfigure}
  \begin{subfigure}{0.95\textwidth}
  \centering
  \includegraphics[width=1\linewidth]{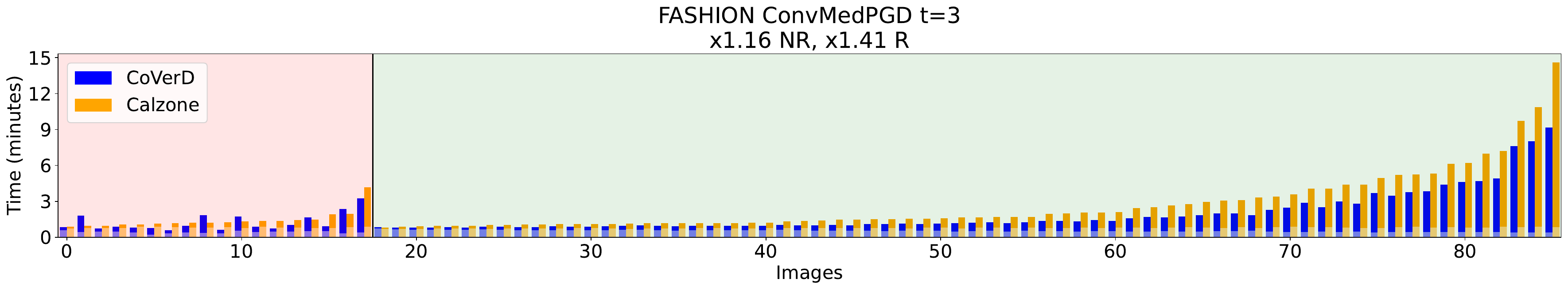}
  \end{subfigure}
  \caption{\tool vs. Calzone for $t=3$.}
  \label{fig:tthre}
\end{figure*} 

\Cref{fig:tthre} compares \tool and Calzone for different networks and $t=3$.

\begin{table}[t!]
\small
\begin{center}
\caption{The average number of completed analysis tasks of GPUPoly and the MILP verifier, 
for robust $t$-balls (R) and non-robust $t$-balls (NR).}
\begin{tabular}{lllcccc}
    \toprule
Dataset & Network\hspace{1.5cm} & $t$ \hspace{1.5cm}& \multicolumn{2}{c}{GPUPoly} & \multicolumn{2}{c}{MILP}\\
\cmidrule(lr){4-5} \cmidrule(lr){6-7}
& & & R & NR & R & NR\\

\midrule
MNIST
 & 6$\times$200\_PGD & 3 & 46565.93 & 12939.12 & 2.99 & 34.88\\
 &  & 4 &  3468151.25 & 637245.50 & 363.00 &  3567.00\\ \cmidrule{2-7}
 & ConvSmall & 3 & 20693.16 & 5822.55 & 0.03 & 2.00\\
 &  & 4 & 615418.88 & 221191.00 & 0.03 & 8.17\\
 &  & 5 & 3095356.00 & 36720.00 & 0.00 & 18.00\\ \cmidrule{2-7}
 & ConvSmallPGD & 3 & 9937.48 & 7689.20 & 0.01 & 3.50\\
 &  & 4 & 171847.55 & 13430.30 & 0.00 & 3.90\\
 &  & 5 & 1730609.17 & 35571.33 & 0.00 & 1.67\\
 &  & 6 & 5856433.00 & 29218.33 & 0.00 & 3.67\\\cmidrule{2-7}
 & ConvMedPGD & 3 & 10617.40 & 11101.56 & 0.01 & 2.67\\
 &  & 4 & 274938.91 & 59813.17 & 0.05 & 4.33\\
 &  & 5 & 3773350.00 & 13614.00 & 0.00 & 5.00\\ \cmidrule{2-7}
 & ConvBig & 3 & 217192.22 & - & 2.33 & -\\
\midrule
FASHION
 &  ConvSmallPGD & 3 & 25196.52 & 15810.41 & 1.76 & 48.65\\
 &   & 4 & 644109.33 & 66506.73 & 1.80 & 67.45\\
 &   & 5 & 2714415.20 & 976522.50 & 33.00 & 8.00\\ \cmidrule{2-7}
 &  ConvMedPGD & 3 & 69537.46 & 42235.61 & 8.25 & 35.33\\
 &   & 4 & 1560073.43 & 60455.00 & 0.00 & 58.00\\
\midrule
CIFAR-10
 &  ConvSmallPGD & 3 & 5768227.40 & 92157.75 & 1933.60 & 1509.75\\
\bottomrule
\end{tabular}
    \label{tab:l_inf_calls}
\quad
\end{center}
\end{table}

We next study the number of completed analysis tasks performed by GPUPoly and the MILP verifier.
For robust $t$-balls, this number is equal to the number of tasks \tool submits to these verifiers. For non-robust $t$-balls, this number may be smaller than the number of submitted tasks, since \tool terminates when an adversarial example is found. 
We note that \tool, like Calzone, submits to the MILP verifier unique blocks (of size $t$) to avoid repeating the same analysis tasks in the MILP verifier, which is significantly slower than GPUPoly.          
\Cref{tab:l_inf_calls} shows the average number of completed analysis tasks for robust $t$-balls and non-robust $t$-balls. 
As expected, the number of completed tasks of the MILP verifier is significantly smaller than the number of completed tasks of GPUPoly, especially for robust balls. This is because 
the MILP verifier is executed for \emph{unique} blocks of \emph{size $t$} that GPUPoly \emph{failed} to verify.

Next, we justify the value chosen for $\text{MAX\_K}$,  
the upper bound on the size of blocks in the \cvd candidates. 
This hyper-parameter has been introduced by Calzone to limit the $k$ of the considered coverings. 
Limiting the maximal value of $k$ is required for the covering database, the sampling, and the dynamic programming. 
This limitation should not impact Calzone's performance, because overly large blocks are unlikely to be robust and thus are unhelpful to the analysis. In Calzone, $\text{MAX\_K}=99$ since for most networks and images, larger blocks are typically not robust. To illustrate, \Cref{fig:successrate} shows the success rate of GPUPoly for 
the ConvSmallPGD network trained for MNIST and neighborhoods $I_S(x)$ as a function of $k=|S|$, for eight values of $x$ (the first eight images in the MNIST test set).
We note that ConvSmallPGD has the highest success rates for large values of $k$, among our evaluated networks. 
For every image, we compute the success rate of every $k\in [200]$ by sampling $5,000$ sets $S$ of size $k$ and submitting $I_S(x)$ to GPUPoly. 
The success rate is the fraction of verified sets. The plot shows that the success rates are very low for $k>99$, justifying Calzone's choice of \text{MAX\_K}.
Unlike Calzone, \tool computes for the first covering a \cvd, consisting of blocks of different sizes. 
A \cvd whose mean block size is $k$ contains blocks larger than $k$. To allow \tool to consider \cvds with mean block sizes similar to those in Calzone's coverings, we increase $\text{MAX\_K}$ to $200$. This number enables \tool to consider \cvds with mean block size $k \leq 99$. This follows since, by \Cref{mean_variance_thm}, the standard deviation of the block size distribution of a \cvd with mean block size $k \leq 99$ is at most $\sqrt{99} \leq 10$. Thus, the probability that a \cvd with mean $k\leq 99$ has a block of size larger than $200$ is negligible, since such sizes are greater than the mean by at least $10$x the standard deviation.      

\begin{figure*}[t]
    \centering
  \includegraphics[width=0.49\linewidth, clip]{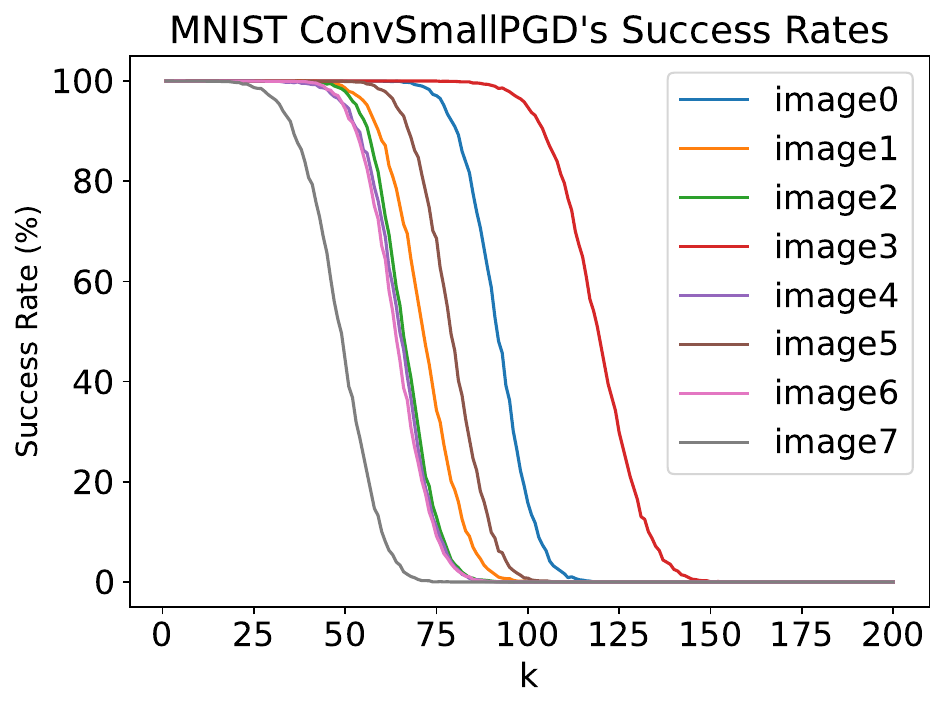}
    \caption{GPUPoly's success rate for MNIST ConvSmallPGD and $I_S(x)$ as a function of $k=|S|$, for eight values (images) of $x$.}
    \label{fig:successrate}
\end{figure*}

Lastly, we justify the value of $n_{\text{fail}}=10$. As described, \tool's sampling is different from Calzone's sampling. Calzone samples $n_\text{samples}=400$ sets of pixels of size $k$ for every $ t\leq k \leq 99$. \tool begins by sampling $n_\text{samples}=400$ sets of pixels of size $k$, for every $ t\leq k \leq 200$ and decreases $n_\text{samples}$ to $24$ when detecting $n_\text{fail}$ values of $k$ whose estimated success rate is zero. Technically, the sampling is distributed among the eight GPUs, each samples independently. Each GPU samples $n_\text{GPU\_samples}=50$ sets of size $k$ for every $k$ from $t$ to $200$, in an increasing order. After a GPU observes $n_{\text{fail}}$ values of $k$ whose estimated success rate is zero, it sets $n_\text{GPU\_samples}=3$. By decreasing $n_\text{GPU\_samples}$, the sampling time decreases while still enabling \tool to estimate the success rate and average analysis time of large values of $k$. 
We evaluate \tool's sampling over 
a single image and four networks. 
\Cref{tab:sampling} shows the minimal $k$ where at least one GPU decreases $n_\text{GPU\_samples}$, denoted $k_{\text{start}}$, and the minimum $k$ where all GPUs decrease $n_\text{GPU\_samples}$, denoted $k_{\text{all}}$. 
We compare \tool's estimated success rates to the estimation of Calzone's sampling extended to $k\leq 200$,
relying on $400$ samples for every $k$. 
For all networks, the estimated success rates of $k \geq k_{\text{start}}$ of \tool and Calzone are zero, except for MNIST ConvSmall and $k=97$, where Calzone's sampling estimated the success rate as $0.25\%$. 
\Cref{fig:sampling_times} shows the estimated analysis time of every $k$ by \tool's sampling and by Calzone's sampling.
The dashed vertical lines in the plots indicate $k_{\text{start}}$. The plots show that the estimated analysis time of \tool is very close to Calzone's estimated analysis time, despite of sampling significantly fewer sets. As 
 expected, its estimations are noisier for $k\geq k_{\text{start}}$. 
 The benefit of \tool's sampling is that it significantly reduces the overall sampling time, on average by $2.65$x and up to $3.39$x compared to Calzone's sampling (extended to $k\leq 200$).

\begin{table}[t!]
\small
\begin{center}
\caption{The value of $k$ when at least one GPU decreases $n_\text{samples}$ and when all GPU decrease $n_\text{samples}$.}
\begin{tabular}{llll}
    \toprule
Dataset & Network\hspace{1.5cm} & $k_{\text{start}}$ & $k_{\text{all}}$ \\
\midrule
MNIST
 & 6$\times$200\_PGD & 69 & 72\\ \cmidrule{2-4}
 & ConvSmall & 94 & 99 \\ \cmidrule{2-4}
 & ConvMedPGD & 90 & 94\\ 
\midrule
FASHION
 &  ConvMedPGD & 62 & 65\\
\bottomrule
\end{tabular}
    \label{tab:sampling}
\quad
\end{center}
\end{table}

 \begin{figure*}[h]
\centering 
\begin{subfigure}[t]{.49\textwidth}
 \centering
\includegraphics[width=1\linewidth]{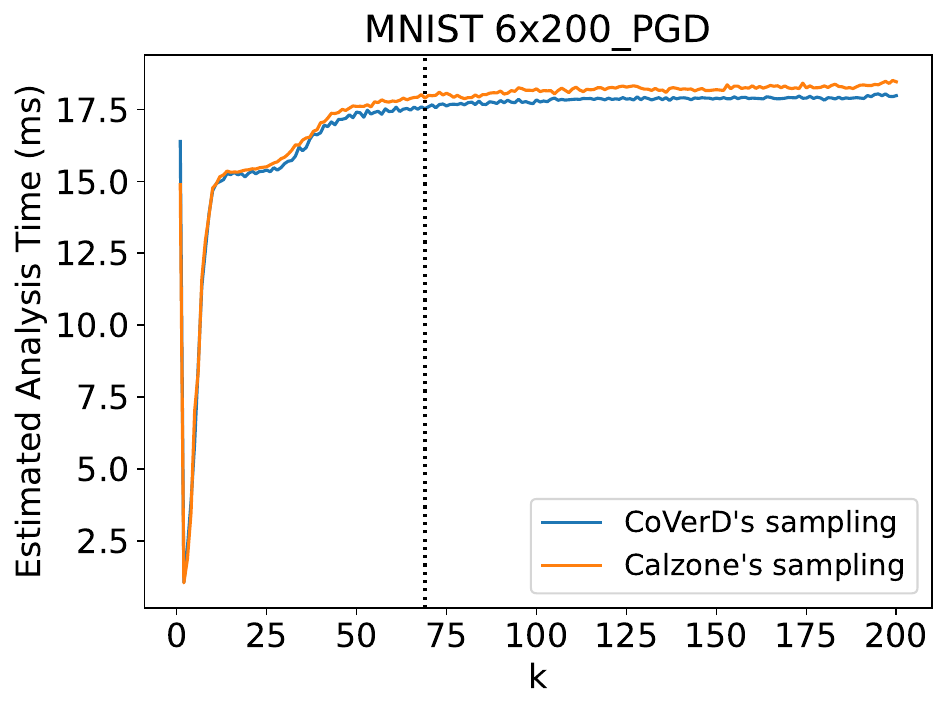}
\end{subfigure}%
\begin{subfigure}[t]{.49\textwidth}
 \centering
\includegraphics[width=1\linewidth]{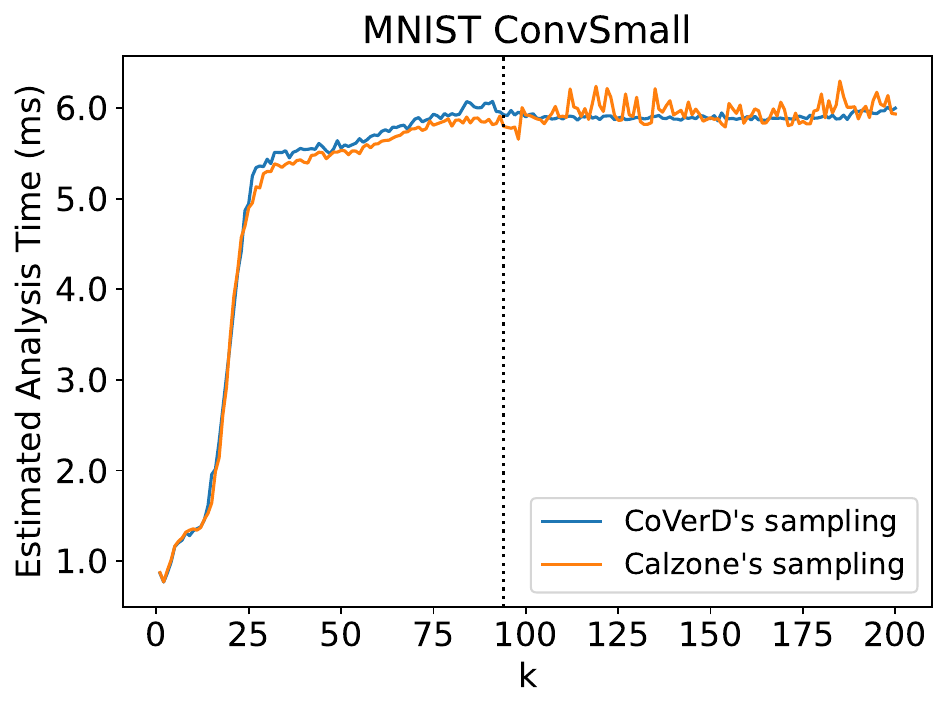}
\end{subfigure}
\begin{subfigure}[t]{.49\textwidth}
 \centering
\includegraphics[width=1\linewidth]{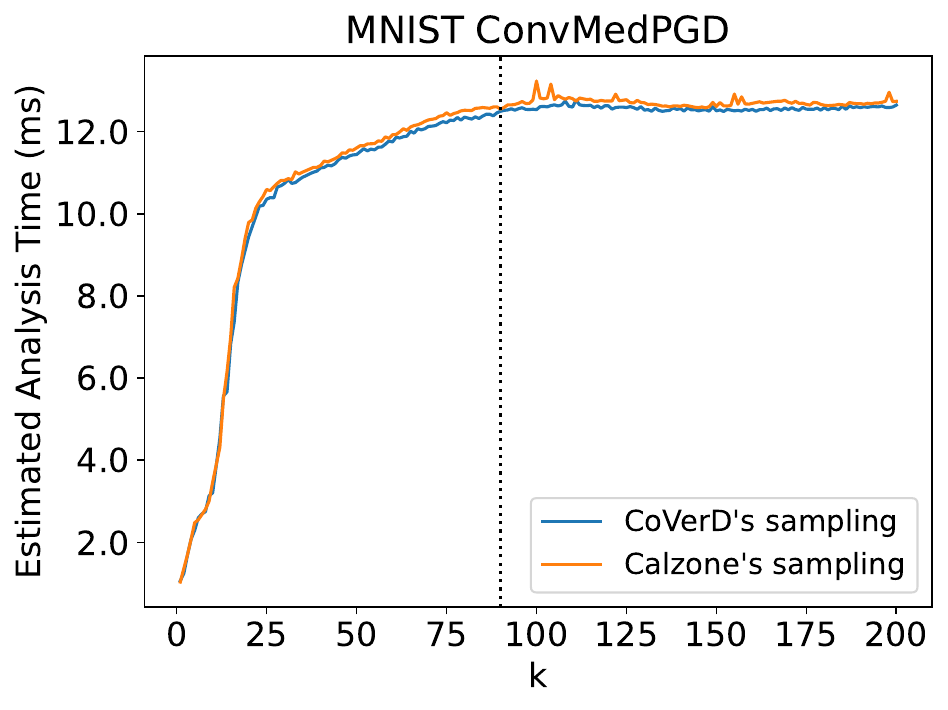}
\end{subfigure}%
\begin{subfigure}[t]{.49\textwidth}
 \centering
\includegraphics[width=1\linewidth]{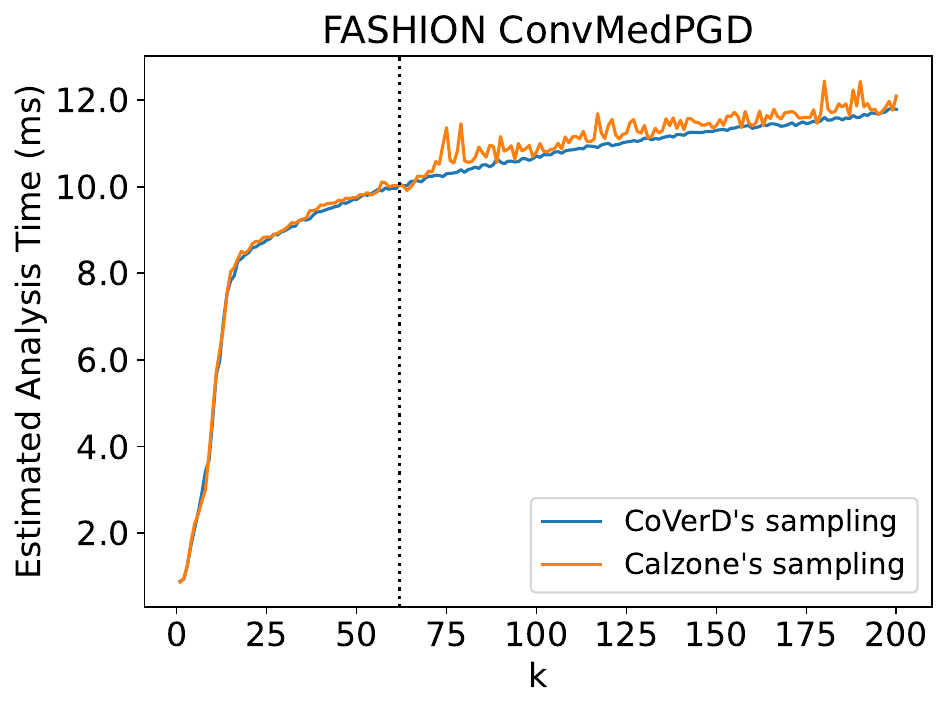}
\end{subfigure}
\caption{Estimated analysis time by \tool's sampling vs. Calzone's sampling.}
    \label{fig:sampling_times}
\end{figure*}
}
\end{document}